\documentclass[11pt,dvipdfmx]{article} %dvipdfmx

\usepackage[truedimen,margin=25truemm]{geometry}

\usepackage{algorithm}
\usepackage{algorithmic}

\usepackage{color}

\usepackage{graphicx} 
\usepackage{epsfig}
\usepackage{subfigure}
\graphicspath{%
{./figures/}
}

\newcommand{\argmax}{\mathop{\rm arg~max}\limits}

\usepackage{amssymb}
\usepackage{amsmath}
\usepackage{amsthm}
\usepackage{mathtools}
\usepackage[raggedright]{titlesec} % avoid hyphen in sections

\def\vec#1{\mbox{\boldmath $#1$}}
\def\mat#1{\mbox{\bf #1}}
\newcommand{\Emph}[1]{\it{#1}}

\newcommand{\Eqref}[1]{(\ref{#1})}
\newcommand{\Figref}[1]{Figure~\ref{#1}}
\newcommand{\Secref}[1]{Section~\ref{#1}}
\newcommand{\Tabref}[1]{Table~\ref{#1}}
\newcommand{\Algref}[1]{Algorithm~\ref{#1}}

\usepackage{amsmath}
\usepackage{amsthm}

\title{State Duration and Interval Modeling in \\Hidden Semi-Markov Model for Sequential Data Analysis}

\author{Hiromi Narimatsu\thanks{H. Narimatsu is with the Graduate School of Information Systems, The University of Electro-Communications, 1-5-1 Chofugaoka, Chofu-shi, Tokyo, 182-8585, Japan (e-mail: narimatsu@appnet.is.uec.ac.jp)} \and Hiroyuki Kasai\thanks{H. Kasai is with the Graduate School of Informatics and Engineering, The University of Electro-Communications, 1-5-1 Chofugaoka, Chofu-shi, Tokyo, 182-8585, Japan (e-mail: kasai@is.uec.ac.jp)}}

\begin{document}

\maketitle

\begin{abstract}
Sequential data modeling and analysis have become indispensable tools for analyzing sequential data, such as time-series data, because larger amounts of sensed event data have become available. These methods capture the sequential structure of data of interest, such as input-output relations and correlation among datasets. However, because most studies in this area are specialized or limited to their respective applications, rigorous requirement analysis of such models has not been undertaken from a general perspective. Therefore, we particularly examine the structure of sequential data, and extract the necessity of ``state duration" and ``state interval" of events for efficient and rich representation of sequential data. Specifically addressing the hidden semi-Markov model (HSMM) that represents such state duration inside a model, we attempt to add representational capability of a state interval of events onto HSMM. To this end, we propose two extended models: an interval state hidden semi-Markov model (IS-HSMM) to express the length of a state interval with a special state node designated as ``interval state node"; and an interval length probability hidden semi-Markov model (ILP-HSMM) which represents the length of the state interval with a new probabilistic parameter ``interval length probability." Exhaustive simulations have revealed superior performance of the proposed models in comparison with HSMM. These proposed models are the first reported extensions of HMM to support state interval representation as well as state duration representation.
\end{abstract}

\begin{center}
\vspace*{0.3cm}
\textcolor{blue}{
{\small
Published in Annals of Mathematics and Artificial Intelligence  \cite{Narimatsu_AMAI_2017}
}
}
\end{center}
\vspace*{0.3cm}

% 
% 
% 
% 
%%%%%%%%%%%%%%%%%%%%%%%%%%%%%%%%%%
% 
%  I.	Introduction
%%%%%%%%%%%%%%%%%%%%%%%%%%%%%%%%%%

\section{Introduction}
\label{sec:Introduction}

The {remarkable} progress of portable devices and wearable devices with multi-functional sensors has enabled {people} to record all the sensing data {easily} and to record all observed events and phenomena. {These circumstances} motivate {people} to analyze such recorded data{. M}any studies have explored {widely diverse} methods of pattern recognition, biological data analysis, speech recognition, image classification, behavior recognition, and time-series data analysis. Esmaeili {\it et al.} categorized {sequential patterns of three types} after theoretical investigation for a large amount of data \cite{Esmaeili2010}. 
Lewis {\it et al.} proposed a sequential algorithm using queries to train text classifiers \cite{Lewis1994}. Song {\it et al.} proposed a sequential clustering algorithm for gene data \cite{Song2009}. More recently, studies using sensor data analysis for human behavior recognition and video sequence understanding have received {considerable} attention because of the {remarkable} progress on wearable devices and the {wider use} of video surveillance systems \cite{Banaee2013,Zheng2014,Cheng2013}. 
Those devices enable users to record all of their experiences such as what is viewed, what is heard, and what is noticed. Nevertheless, although collecting all observed data has become much easier, it remains difficult to find data that we want to access {immediately} because the amount of time-series data is extremely huge. In {the} case of {\it life log} data application, for example, it must be easy to retrieve information {related to} particular places or dates if rich and comprehensive {\it meta-data} are attached {sufficiently} to every datum to be identified. However, if a query is ambiguous{, such as} retrieving {\it a situation similar to the current situation where 10-minute continuous ``Event A" starts 30 minutes later after half-hourly ``Event B" finishes}, {then} it must {surely be} challenging to obtain meaningful results. {Consequently}, finding such {\it similar sequential patterns} from vast sequential data using a given target pattern extracted from the current situation is of crucial importance. This is {a point of} interest {examined in this study}. Finding similar sequential patterns {requires discrimination of} particular sequential patterns from many {\it partial groups of multiple events} of patterns. For this purpose, among the specialized methods {used} to detect similar partial patterns from sequential data which include, for instance, Dynamic Programming (DP) matching algorithm, and Support Vector Machine (SVM), this {study specifically examines a} hidden Markov model (HMM) because HMM is specialized and promising to {address} sequential data by exploiting transition probability between states, i.e., events.

The primary contributions of our work are two-fold: (a) we advocate that the support of both ``state duration" and ``state interval" is of great {importance} to represent practical sequential data based on studies about the feature{s} and structure{s} of sequential data{. T}hen we extract requirements for its modeling. Next, (b) we propose two sequential models by extending hidden semi-Markov model (HSMM) \cite{Yu2010} to support both {the} state duration and {the} interval of events efficiently. More concretely, regarding (a), we especially address the generalization of model requirements for sequential data, and emphasize the importance of handling {the} event order, continuous time length, i.e., state duration of an event, and discontinuous time length, i.e., state interval between two events. {This report} is the first {of the relevant literature describing} generalization of the model requirements for sequential data. Herein, we define the continuous duration time of a state as {the} state duration, and define the discontinuous interval with no observation as {the} state interval because an event is treated as a state in HMM. Then, with respect to (b), after assessment of the extended HMM models in the literature against those requirements, we show that {none of} the existing models {treats} both {the} state duration and {the} state interval simultaneously. Nevertheless, we also show that HSMM, {an} extended HMM model, handles state duration, and {that it} is an appropriate baseline to be extended to meet all demands. Subsequently, {this report proposes} two extended models by extending HSMM that accommodates not only {the} state duration but also {the} state interval.

Two approaches are specifically addressed to treat {the} state interval with HSMM. For both approaches, three variations can be {regarded as the} model {representing} {the} state interval{:} modeling state interval with (i) only {a} preceding state, (ii) only {a} subsequent state, and (iii) both preceding and subsequent states. From the viewpoint of modeling accuracy, we specifically examine modeling of {the} state interval with both preceding and subsequent states. {Finally}, we propose two extended models of HSMM: one represents {the} state interval as a new {\it node of state interval}{;} the other represents {the} state interval by a new {\Emph{probability of state interval length}}. The first model, dubbed {the} interval state hidden semi-Markov model (IS-HSMM), is categorized into a straightforward extension of the original HSMM. The distinct difference is the introduction of a new ``interval state node." {S}imple introduction of the interval state node into HSMM, however, {engenders} improper transition probabilities because the transition frequencies {of the} {general} state to the new interval state and the transition from the interval state to another state {might} increase when the {interval state} symbols are observed frequently. This causes undesired biases onto the original transition probability, and finally brings severe degradation of model accuracy. To {re}solve this issue, IS-HSMM expresses {a} {second-order} Markov model at the part where the preceding state is the interval state node. {T}he second proposed model {is designated} as {the} interval length probability hidden semi-Markov model (ILP-HSMM){,} {and it} represents {the} state interval by a new parameter to HSMM. This parameter is {the} ``interval length probability{,}" which is represented as a probability density distribution function, and {which} is modeled with the two combined states. Preliminary studies of ILP-HSMM {were} proposed in our earlier work as DI-HMM \cite{Narimatsu2015}.

The remainder of this paper is organized as follows. The next section introduces related {work}. \Secref{sec:SequentialDataAnalysis} presents a description of the model requirements and {a} requirement assessment of the existing HMM variants. Then, a brief explanation of the original HMM model is given. \Secref{sec:HSMM} explains the baseline model of our proposal: {a} hidden semi-Markov model, i.e., HSMM. After examining the approaches for state interval modeling based on HSMM in \Secref{sec:ModelingDurationAndIntervalInHSMM}, we propose {the} two models, IS-HSMM and ILP-HSMM, respectively, in \Secref{sec:HSMMwithIntervalState} and \Secref{sec:ILPHSMM}. Finally, we {demonstrate the} superior performance of the proposed models in comparison with HSMM in \Secref{sec:Evaluation}. We summarize the results presented in this paper and describe {avenues of} future work in \Secref{sec:Summary}.

% 
% 
% 
% 
%%%%%%%%%%%%%%%%%%%%%%%%%%%%%%%%%%
% 
%  II.	Related Work
%%%%%%%%%%%%%%%%%%%%%%%%%%%%%%%%%%
\section{Related {Work}}
\label{sec:RelatedWork}

This section presents related work {that has been reported in the field of} sequential data analysis. For sequential pattern matching and sequential pattern detection, the Dynamic Programming (DP) algorithm \cite{Sakoe1971} provides an optimized search algorithm that calculates the cost of a path in a grid and which thereby finds the least costly path. {Actually, DP} was first used for acoustic speech recognition. For sequential pattern classification, Support Vector Machine (SVM) \cite{Vapnik1998,Abe2010} is a classifier that converts {an} $n$-class problem into multiple two-class problems. SVM has {demonstrated} its superior performance in a {diverse} applications such as face and object recognition from a picture. {Regarding the} Regression Model (RM) \cite{Boscardin1996}, the logistic regression model \cite{Cox1958} is a representative model that is powerful binary classification model when the model parameters are {mutually} independent. {The h}idden Markov model (HMM), originally proposed in \cite{Baum1966,Baum1967}, is a statistical tool used for modeling generative sequences. HMM has been {used} frequently together with the Viterbi algorithm to estimate the likelihood of generating observation sequences. Whereas HMM is {used} widely for many applications {such as} speech recognition, handwriting recognition{,} and activity recognition, many extended HMMs have {also been} proposed to enhance the expressive capabilities of the baseline HMM model {and} to support various specialized application data. {Concequently}, addressing HMM {as} a powerful and robust model for treating sequential data using its transition probability in a statistical manner, we particularly examine HMM in the present paper.

With regard to the extensions of HMM, Xue {\it et al.} proposed transition-emitting HMMs (TE-HMMs) and state-emitting HMMs (SE-HMMs) to treat the discontinuous symbol \cite{Xue2006}, of which application is an off-line handwriting word recognition. The observation data {include} discontinuous and continuous symbols between characters when writing in cursive letters. They specifically examined such discontinuous features and continuous features, and extended HMM to treat {both}. Bengio {\it et al.} {specifically examined} mapping {of} input sequences to the output sequences \cite{Bengio1996}. The proposed model supports a recurrent networks processing style and describes an extended architecture under the supervised learning paradigm. Salzenstein {\it et al.} dealt with a statistical model based on Fuzzy Markov random chains for image segmentations in the context of stationary and non-stationary data \cite{Salzenstein2007}. They specifically examined the observation in a non-stationary context, and proposed a model and a method to estimate model parameters. {Ferguson proposed a variable duration models of HMM for speech recognition{. Today,} the model is familiar {as} the extended model of HMM as explicit-duration hidden Markov model or hidden semi-Markov model \cite{Yu2003,Mitchell1993,Ramesh1992,MurphyThesis2002}. They proposed a new forward{-}backward algorithm to estimate model parameters.}

Addressing the difference of duration in each state, hidden semi-Markov {m}odel (HSMM) is proposed to treat the duration and multiple observations produced in {a} single state \cite{Yu2010,Murphy2002}. 
{The {salient} difference between HMM and HSMM is whether it can treat the duration of states in HMM{. The} technique of EM algorithm{s} for modeling the duration of states was proposed by Ferguson \cite{Ferguson1980}. He proposed the algorithm for speech recognition, {but} the model is further applied for time-series data {for} word recognition {and} rainfall data \cite{Guedon1990,Sansom1999,Sansom2001,Guedon2003}. Then,} 
Bulla proposed an estimation procedure to the right-censored HSMM for modeling financial time-series data using conditional Gaussian distributions for the HSMM parameters \cite{Bulla2006,Bulla2006b}. For diagnosis and prognosis using multi-sensor equipment, Dong {\it et al.} prioritized the weights for each sensor to treat multiple sensor results, and showed that the proposed model of HSMM gave higher performance than the original HSMM \cite{Dong2006}. Recently, Dasu analyzed HSMM and described how to implement HSMM for a practical application in detail \cite{Dasu2011}. 
{Baratchi {\it et al.} and Yu {\it et al.} proposed {extended} hidden semi-Markov models for mobility data.  \cite{Yu2003MobilityTracking,Baratchi2014} These models can treat the sequential data which include missing data.}

%%%%%%%%%%%%%%%%%%%%%%%%%%%%%%%%%%
% 
% III.	Sequential Data Analysis
%%%%%%%%%%%%%%%%%%%%%%%%%%%%%%%%%%
\section{Analysis of Sequential Data Modeling}
\label{sec:SequentialDataAnalysis}
This section presents an analysis of sequential data modeling and derives the model requirements for sequential data analysis. Then, the satisfactions of the extended models of HMM for the model requirements are examined.
%
%%%%%%%%%%%%%%%%%%%%%%%%%%%%%%%%%%
% 
% III-A.	Requirement for Model Description
%%%%%%%%%%%%%%%%%%%%%%%%%%%%%%%%%%
\subsection{Requirement for Model Description}
\label{sec:RequirementForModel}
% 
% 
%
%%%%%%%%%%%%%%%%%%%%%%%
% Figure 1
%%%%%%%%%%%%%%%%%%%%%%%
\begin{figure}[tb]%htbp
\begin{center}
\includegraphics[width=0.95\columnwidth]{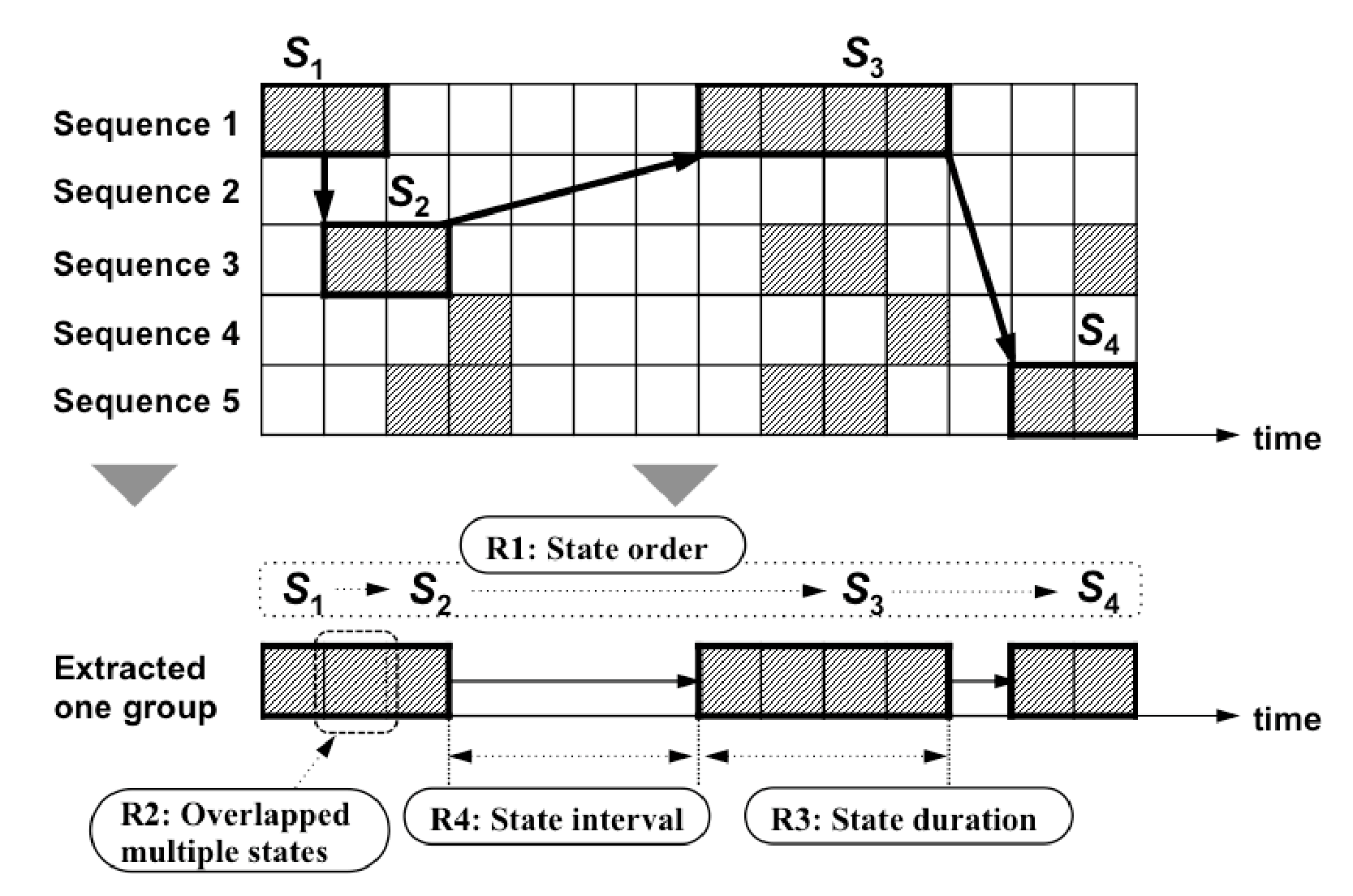}
\caption{Event generative model and sequential data model requirements.}
\label{fig:RequirementForTheModel}
\end{center}
\end{figure}
This section presents discussions {of} the requirements for model description using time-series data: representative data of sequential data. For this purpose, we assume {a} situation {in which} multiple different sequences are {generated} independently from five sensors as shown in \Figref{fig:RequirementForTheModel}. Here, an observed event of which value {of the sensor} exceeds a predefined threshold is recognized as a `state' represented in a block. The continuous period of each event is represented by the {block length}. {Because} events are not successively observed, {a} {\it no-observation period} exists between two successive states in certain periods{. T}he length of such {a} no observation period is represented as the distance between two blocks. 
In this example, we also assume that a set of four black blocks, \{$S_1$, $S_2$, $S_3$, $S_4$\}, express{es} an extracted multiple states that forms one particular group.

Now we extract the requirements for model description. First, addressing this formation of four blocks, {it is readily apparent} that these states are observed in a prescribed order. {Therefore, it is apparent} that the order of multiple states should be described in a model ({\bf R1}). 
Second, multiple states {are visible} in a partially overlapped manner{,} as {shown by} $S_1$ and $S_2$. In other words, multiple states can occur simultaneously at a certain period. Therefore, the model {must} support the representation capability to describe multiple states occurring at the same time ({\bf R2}). Third, {because} the time lengths of respective states mutually differ, the {state} duration {must} be expressed in a model ({\bf R3}). Finally, for the case in which each state occurs intermittently, a %vacant 
{no observation} period between one state and another state that is not involved in the group of sequence might exist between two states. Furthermore, the length of this %vacant 
{no observation} period shall be variable. Therefore, {the} state interval between two states in a model must be described ({\bf R4}). In summary, the sequential data model is required to describe these requirements{. This report} defines these respective requirements as follows{.}

\begin{enumerate}
\renewcommand{\labelenumi}{(\roman{enumi})}
 \item {\bf R1:} State order
 \item {\bf R2:} Staying multiple states in a certain period
 \item {\bf R3:} State duration
 \item {\bf R4:} State interval
\end{enumerate}
\vspace*{0.3cm}

Among these items, {\bf R2} differs from other items because {\bf R1}, {\bf R3}, and {\bf R4} are required even for a single sequence{, whereas} {\bf R2} is the requirement for multiple sequences. Therefore, this study specifically examines requirements {\bf R1}, {\bf R3}, and {\bf R4}. The examination of {\bf R2} shall be left for advanced {studies} to be undertaken {as} future work.

%%%%%%%%%%%%%%%%%%%%%%%%%%%%%%%%%%
% 
% III-B.	Requirement Verification for Extended HMM Methods
%%%%%%%%%%%%%%%%%%%%%%%%%%%%%%%%%%
\subsection{Requirement Verification for Extended HMM Models}
\label{sec:RequirementVerification}

%%%%%%%%%%%%%%%%%%%%%%%
% Table 1 tab:Satisfaction  %els
%%%%%%%%%%%%%%%%%%%%%%%
\begin{table}[tb]%h
%\footnotesize
\begin{center}
 \caption{Requirement satisfactions in HMM, HMM variants, and our proposals.}
 \begin{tabular}{l|c|c}\hline 
 \shortstack{}&\multicolumn{2}{|c}{Requirements}\\ %\hline
 \cline{2-3}
% \shortstack{Model}&\shortstack{\ \ ({\bf R3}) \ \ }&\shortstack{({\bf R4}) }\\ 
 \multicolumn{1}{c|}{Model}&\shortstack{\ \ Time length\ \ }&\shortstack{ Time Interval}\\ 
 \shortstack{}&\hspace*{0.5cm}\shortstack{in a state ({\bf R3})}\hspace*{0.5cm}&\shortstack{between states ({\bf R4})}\\ \hline 
 \hline 
 HMM (baseline) \cite{Eddy1995}&&\\ \hline
 HMM-selftrans \cite{Xue2006} &\checkmark&\\ \hline
 FO-HMM \cite{Salzenstein2007}&&\\ \hline
 IO-HMM \cite{Bengio1996}&&\\ \hline
EDM \cite{Yu2003,Mitchell1993} and HSMM \cite{Yu2010,Murphy2002}&\checkmark&\\ \hline
{{\bf IS-HSMM} and {\bf ILP-HSMM} (proposal)}&\checkmark&\checkmark\\ \hline
 \end{tabular}
 \label{tab:Satisfaction}
\end{center}
\end{table}

This section {presents investigation of} whether HMM and the extended variants of HMM satisfy those requirements. \Tabref{tab:Satisfaction} {presents} a comparison among the existing HMM models from the viewpoints of the model requirements described above. {Because} the baseline HMM model describes the order of the states ({\bf R1}), all the extended HMM models inherit this capability. FO-HMM is specialized for treating the ambiguity of observation symbols{. It} does not contribute to our model requirement. IO-HMM is a hybrid model of generative and discriminative models to treat the estimation probability commonly used for input sequence and observations. Therefore, it does not satisfy the remaining requirements. HSMM models the time length to remain in {a} single state \cite{Yu2010}{. I}ts variants including HMM-selftrans and EDM \cite{Yu2003,Mitchell1993,Ramesh1992,MurphyThesis2002} satisfy the same requirements: state order ({\bf R1}) and state duration ({\bf R3}).

As a result of investigation of the requirement satisfaction, {it is apparent that} no existing HMM model accommodates both {the} state duration and {the} state interval together. Nevertheless, we conclude that HSMM is the best baseline model to be extended towards our new target model because only HSMM handles state duration.

{Moreover, some {extended} models of HSMM {have} been proposed. Baratchi {\it et al.} and Yu {\it et al.} proposed {extended} models of HSMM {that} can treat missing data. Their proposal can model the sequential data even {if they} include missing intervals \cite{Yu2003MobilityTracking,Baratchi2014}. These studies are motivated to complement the missing data so {that} the {`}interval of missing' might have variable status in all sequence{s}. It is useful for modeling even if it has missing data and %completing 
{interpolating} the missing data. However{,} in the situation we lead from the sequential data analysis described in this section, the {\it interval} is not {`}missing'. The status of {the} interval is only the interval which includes other status {that is un}related to the sequence. Therefore the target for modeling {differs {from our target}. It is necessary} to model the {\it interval} which is not missing. {Therefore}, the next section {provides} a detailed explanation of HSMM.}

%%%%%%%%%%%%%%%%%%%%%%%%%%%%%%%%%%
% 
% IV.	Hidden Semi-Markov Model
%%%%%%%%%%%%%%%%%%%%%%%%%%%%%%%%%%
\section{Hidden Semi-Markov Model (HSMM)}
\label{sec:HSMM}

HMM has been studied as a powerful model for speech recognition.
The model parameters of HMM consist of the initial state probability, the transition probability between states and the emission probability of observation elements from each state. {T}he model training phase calculates the optimum values of the model parameters. {T}he recognition phase calculates the probabilities that generates {an} observed sequence {for each model}, and {then} selects the {highest probability model as a recognition result.}

{T}he distinguishing feature of HMM is to model the transition probability of every pair of two states. However, the time length to stay in each state {cannot be modeled by HMM, which is} 
{fundamentally necessary} for modeling in some useful applications {such as} online handwriting recognition. 
HSMM{, which} has been proposed to support this time length, has long been studied for some specific applications such as speech recognition and online handwriting recognition. This section, after providing basic notation, {presents details of} the algorithms of the model training and recognition in HSMM.

% 
% 
%%%%%%%%%%%%%%%%%%%%%%%
% Figure 2
%%%%%%%%%%%%%%%%%%%%%%%
%
\begin{figure}[tb]
\begin{center}
    \begin{subfigure}[Model structure of HMM. The state node of HMM emits an observation symbol.]
           {\includegraphics[width=0.95\columnwidth]{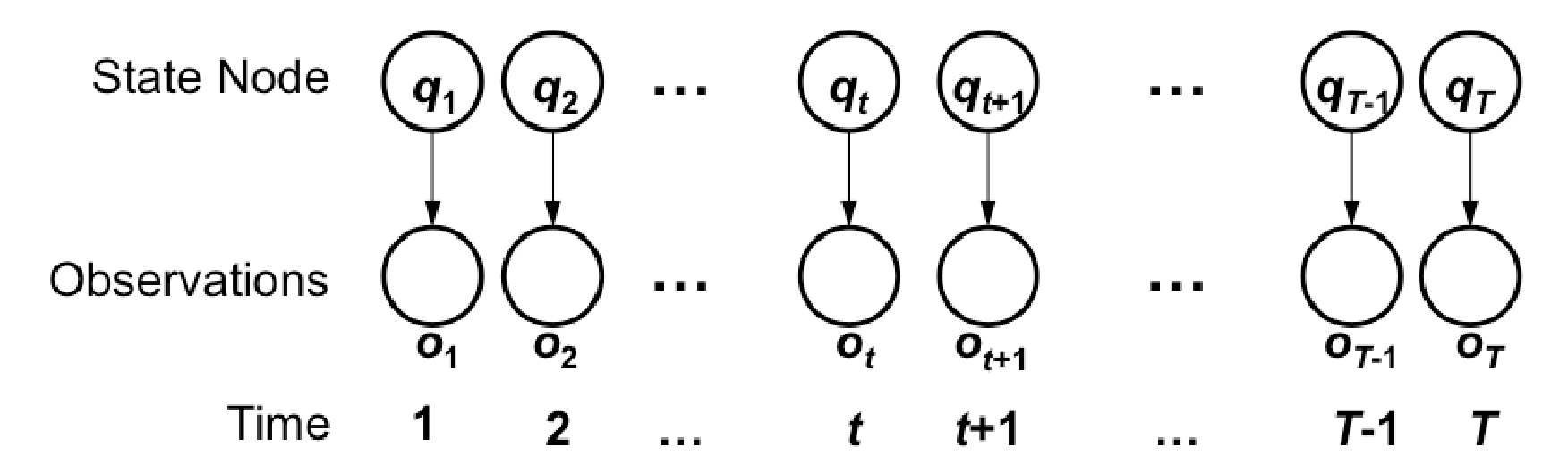}}
    \end{subfigure}
    \begin{subfigure}[Model structure of HSMM. The super state node of HSMM emits observation sequence {in} a certain duration.]
           {\includegraphics[width=0.95\columnwidth]{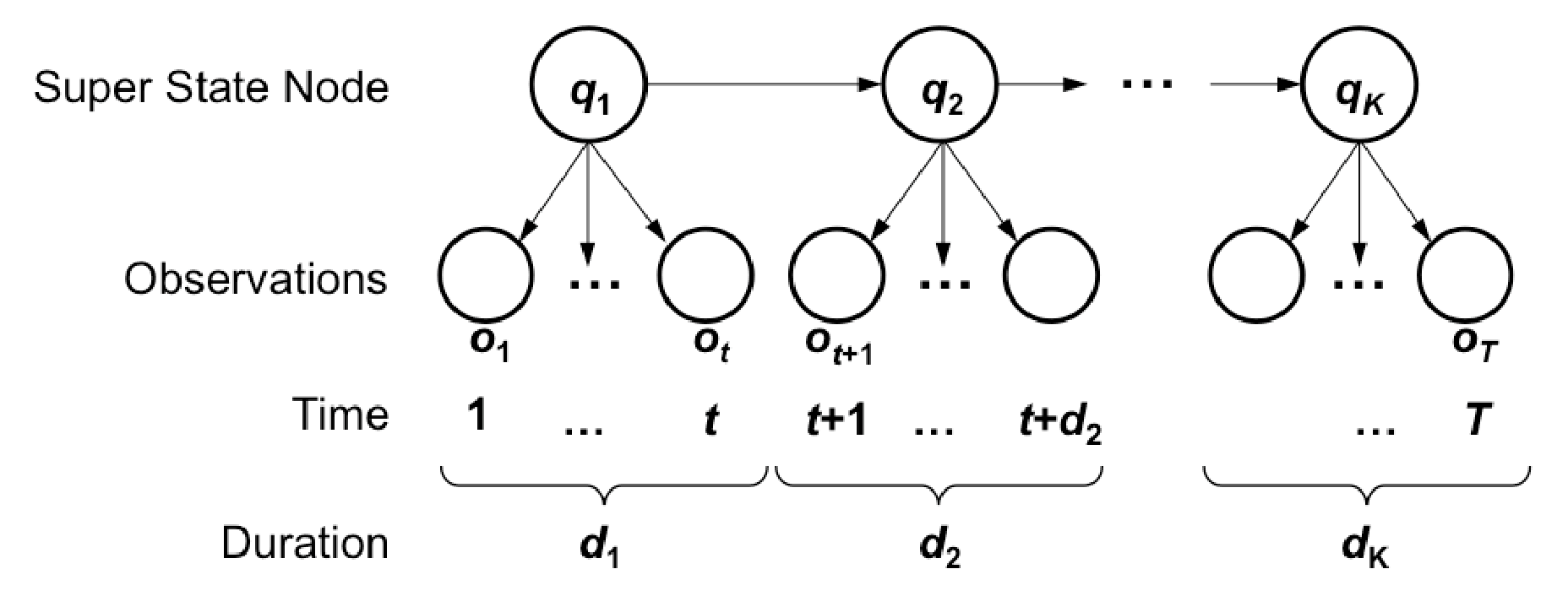}}
    \end{subfigure}
\caption{Model structure comparison between HMM and HSMM.}
\label{fig:HSMM}
\end{center}
\end{figure}
%%%%%%%%%%%%%%%%%%%%%%%
% 

%%%%%%%%%%%%%%%%%%%%%%%%%%%%%%%%%%
% 
% IV-A.	Notation
%%%%%%%%%%%%%%%%%%%%%%%%%%%%%%%%%%
\subsection{Notations}
\label{subsec:Notation}
The HSMM structure is shown in \Figref{fig:HSMM} {compared} with that of HMM. Here{in}after, we assume that each unit time at time $t$ has one corresponding observation $o_{t}$. The observation sequence from time $t={t_1}$ to $t={t_2}$ is denoted as $\vec{o}_{t_1: t_2}=o_{t_1}, ..., o_{t_2}$. A set of output symbols is expressed as $Y=\{y_1, y_2, \cdots ,y_N\}$, where $N$ is the number of {symbols, and $o_t \in Y$.} 
A set of hidden states is $S = \{1, \cdots, M\}$, where $M$ is the number of {hidden states}, and the hidden state sequence from time $t=1$ to $t=T$ is expressed {as} $S_{1:T}=S_1, ..., S_T$, where $S_t$ represents a state at time $t$. 
{Whereas} HMM allows each state node to emit an observation symbol, HSMM {has super{-}state node instead and each super state node can} emit multiple observation symbols, i.e., observation sequence. Here the hidden state sequence is represented as $Q = q_1, \cdots, q_k, \cdots, q_K${,} where $K$ is the number of states in a sequence. {Also,} $K = T$ in HMM, $K \leq T$ in HSMM. 
{T}he $k$-th hidden state {in the sequence }is assigned {to} 
state $i$ as {$q_k = i \in S$} in both HMM and HSMM.

The parameters incorporated in the HMM model are ${\mathrm{\Lambda}} = \{\mat{A}, \mat{B}, \vec{\pi}\}$, where $\mat{A} \in \mathbb{R}^{M \times M}$ is the matrix representing the transition probabilities between states, $\mat{B} \in \mathbb{R}^{M \times N}$ is the matrix for the emission probability from each state, and $\vec{\pi} \in \mathbb{R}^{M}$ represents the initial probability 
{of }each state. 
{T}he transition probability from state $i$ to state $j$ is denoted as $\mat{A}(i,j)=a_{ij}$ where $i,j \in S$. Similarly, the emission probability of symbol $y_n$ from state $j$ is represented as $b_{j}(y_n)$ and \mat{B}$(j,n)=b_{j}(y_{n})${,} 
where $j \in S$ and $y_n \in Y$. The initial probability {that state $i$ occurs} is denoted as {$\pi_i$}.

{However}, HSMM handles the same set of parameters $\mathrm{\Lambda}=\{\mat{A}, \mat{B}, \vec{\pi}\}$, but the elements of each parameter differ from those of HMM to describe the duration of states. The set of duration time{s} is denoted as $D${;} the duration of state $i$ is represented as $d_i \in D$. 
Considering this new parameter, the transition probability from state $i$ to state $j$ is represented as $a_{(i,d_i)(j,d_j)}$ {instead} of $a_{i,j}${. T}he emission probability is represented as $b_{j,d_j}(\vec{o}_{t+1:t+d_j})$ {instead} of $b_j(o_t)$.
{Parameter} $\mathrm{\Lambda}$ is updated by the recursive calculation for inference{. T}he latest calculation result for update is represented as $\hat{\mathrm{\Lambda}}$. 
The overall algorithm is summarized in \Algref{alg:alg1}.

%
%
%%%%%%%%%%%%%%%%%%%%%%%
% pseudocode (HSMM) alg1
%%%%%%%%%%%%%%%%%%%%%%%
\begin{algorithm}[tb]
\caption{Algorithm for training and recognition in HSMM.
}
\label{alg:alg1}                 
\begin{algorithmic}[1] 
\REQUIRE{Input}\\
Training sequences: $\vec{o}_{1:{T_r}}^{z}=\{o_1^z, \cdots, o_{T_r}^z\}$,\\
Testing sequences: $\vec{o}_{1:{T_t}}^{*}=\{o_1^{*}, \cdots, o_{T_t}^{*}\} $. \\
{($Z$ is the number of training sequences.)}\\
{($H$ is the number of recursive calculation.)}\\
\ENSURE {\bf Training phase} \\
\FOR {$z=1$ to $Z$ } 
\STATE{
Assign random values to the HSMM parameters 
$\mathrm{\Lambda}^z = \{\mat{A}, \mat{B}, \mat{$\pi$}\}$, and $\alpha_{t(j,d_j)}$ and $\beta_{t(j,d_j)}$.\\
}
\FOR {$h=1$ to $H$} 
\FOR {$t=1$ to $T_r$} 
\STATE{
  Calculate $\alpha_{t(j,d_j)}$ and $\beta_{t(j,d_j)}$ using \Eqref{eq:eq3} and \Eqref{eq:eq4}.
  }
\STATE{
  Update parameters $\mathrm{\Lambda}^z$ using \Eqref{eq:eq5} and \Eqref{eq:eq6}.
}
\ENDFOR
\STATE{
  Calculate $\theta_h$ using \Eqref{eq:eq7}.
  \IF {$\theta_h - \theta_{h-1} < \epsilon$
  }
  \STATE{
  {\bf break}
  }
  \ENDIF
}
\ENDFOR
\ENDFOR
\ENSURE {\bf Recognition phase}
\FOR {$z=1$ to $Z$}
\FOR {$t=1$ to $T_t$}
\STATE{
  Prepare $\mathrm{\Lambda}^z$ 
  from the results obtained in the training phase.
}
\STATE{
  Calculate $\alpha_t(j,d_j)$
  using \Eqref{eq:eq8}.
}
\ENDFOR
\STATE{
    Calculate $P(o_{1:T_t} | \mathrm{\Lambda}^z)$ using $\alpha_t(j,d_j)$.
}
\ENDFOR
\STATE{
Select the model $z^*$ that has the maximum value for $P(o_{1:T_t}^{*} | \mathrm{\Lambda}^z)$.
\STATE{
{\bf Return} Model $z^*$ and its probability $P(\vec{o}_{1:T_t}^{*} | \mathrm{\Lambda}^{z^*})$.
}
}
\end{algorithmic}
\end{algorithm}

%
%
% 
% 
%%%%%%%%%%%%%%%%%%%%%%%%%%%%%%%%%%
% 
%  IV-B.	Model Training (Inference)
%%%%%%%%%%%%%%%%%%%%%%%%%%%%%%%%%%
\subsection{Model Training (Inference)}
\label{subsec:Training}

This section presents a description of how to train the model of HSMM using t{r}aining sequences, i.e., how to estimate the set of parameters $\mathrm{\Lambda} = \{\mat{A}, \mat{B}, \vec{\pi}\}$ including the duration in each {state}. HSMM {is trained} using Baum{-}Welch algorithm \cite{Baum1966} {in} the same way as HMM, where a recursive forward-backward algorithm is used. The forward{-}backward algorithm is an inference algorithm used for HMM{. A}n extended algorithm special for HSMM is also proposed \cite{Kobayashi2007}.

The concrete algorithm for HSMM is {the following}: computing forward probabilities starts from $t=1$ to $t=T$, {with computed}  backward probabilities from $t=T$ to $t=1$. This two-way calculation repeat{s} until {the likelihood converges. }More concretely, the forward step calculates the following forward variable $\alpha_t(j,d_j)$ of state $j$ with $d_j$ at $t$ as 
%
%%%%%%%%%%%%%%%%%%%%%%%
%  Equation 1 % (4)
%%%%%%%%%%%%%%%%%%%%%%%
%
\begin{eqnarray}\label{eq:eq3}
\alpha_t(j,d_j)&=&\sum_{i \in \{S\} \setminus \{j\}} \sum_{d_{i} \in D}\alpha_{t-{d_j}}(i,d_i) a_{(i,d_i)(j,d_j)} b_{j,d_j}(\vec{o}_{t-d_j+1:t}).
\end{eqnarray}

%
%%%%%%%%%%%%%%%%%%%%%%%
%
The backward step calculates the following backward variable $\beta_t(j,d_j)$ as
%
%%%%%%%%%%%%%%%%%%%%%%%
% Equation 2 %(5) 
%%%%%%%%%%%%%%%%%%%%%%%
%
\begin{eqnarray}\label{eq:eq4}
\beta_t(j,d_j) &=& 
\sum_{i \in \{S\} \setminus \{j\}} \sum_{d_i \in D}
a_{(j,d_j)(i,d_i)} b_{i,d_i}(\vec{o}_{t+1:t+d_i}) \beta_{t+d_i}(i,d_i).
\end{eqnarray}
The calculation step for estimating {the} model parameters are {presented} below.\\
{\bf Step 1 Initialization}
\par Give an initial set of parameters $\mathrm{\Lambda}$ of the model at random.\\
{\bf Step 2 Recursive calculation}
\par 
Calculate the set of parameters $\hat{\mathrm{\Lambda}}$ that maximizes the variables of the forward{-}backward algorithm using the initialized parameter $\mathrm{\Lambda}$. Denoting the updated state transition probability $a$ and the updated emission probability $b$ as $a'$ and $b'$, respectively, $a'_{(i,d_i)(j,d_j)} $ and $b'_{j,d_j}(\vec{o}_{t+1:t+d_j})$ are updated using the previous values of $a_{(i,d_i)(j,d_j)}$ and $b_{j,d_j}(\vec{o}_{t+1:t+d_j})$. More specifically, the state transition probability from state $i$ with $d_i$ to state $j$ with $d_j$ is defined as
% 
% 
%
%%%%%%%%%%%%%%%%%%%%%%%
% Equation 3  %(2)
%%%%%%%%%%%%%%%%%%%%%%%
%
\begin{eqnarray}\label{eq:eq1}
a_{(i,d_i)(j,d_j)} &:=& P(S_{t+1:t+d_j}=j | S_{t-d_i+1:t}=i). \nonumber 
\end{eqnarray}
%
%%%%%%%%%%%%%%%%%%%%%%%
%
Analogous to the state transition probability, the emission probability of $\vec{o}_{t+1:t+d_j}$ from state $j$ with $d_j$ is defined as 
%
%%%%%%%%%%%%%%%%%%%%%%%
% Equation 4  %(3)
%%%%%%%%%%%%%%%%%%%%%%%
%
\begin{eqnarray}\label{eq:eq2}
b_{j,d_j}(\vec{o}_{t+1:t+d_j}) &:=& P(\vec{o}_{t+1:t+d_j} | S_{t+1:t+d_j}=j).
\nonumber
\end{eqnarray}
Then, these probability updates are calculated as \Eqref{eq:eq5} and \Eqref{eq:eq6} using the variables of \Eqref{eq:eq3} and \Eqref{eq:eq4} as
%
%%%%%%%%%%%%%%%%%%%%%%%
% Equation 5 %(6)
%%%%%%%%%%%%%%%%%%%%%%%
%
%% revised 20170326
\begin{eqnarray}\label{eq:eq5}
a'_{(i,d_i)(j,d_j)}&=&
{\frac 
{\sum_{t=1}^{{T}} \alpha_t(i,d_i) a_{(i,d_i)(j,d_j)} {b_{i,d_i}(\vec{o}_{t-d_i+1:t})} \beta_{t+{d_j}}(j,d_j)} 
{\sum_{t=1}^{{T}} \alpha_t(i,d_i) \beta_t(i,d_i)}
}.
\end{eqnarray}
% 
%
%%%%%%%%%%%%%%%%%%%%%%%
%
%%%%%%%%%%%%%%%%%%%%%%%
% Equation 6  %(7)
%%%%%%%%%%%%%%%%%%%%%%%
%
\begin{eqnarray}\label{eq:eq6}
b'_{j,d_j}(\vec{o}_{t+1:t+d_j}) &=& 
{\frac 
{\sum_{t=1}^{T} \delta(o_t,y_n) \alpha_t(j,d_j) \beta_t(j,d_j)} 
{\sum_{t=1}^{T} \alpha_t(j,d_j) \beta_t(j,d_j)}
},
\end{eqnarray}
%
%%%%%%%%%%%%%%%%%%%%%%%
%
%
where $\delta(o_t,y_n)$ is defined as
% 
%
%%%%%%%%%%%%%%%%%%%%%%%
% Equation7'
%%%%%%%%%%%%%%%%%%%%%%%
%
\begin{eqnarray}
\label{eq:eq6_2}
\delta(o_t,y_n) &=& \left\{
\begin{array}{lll}
1 & & {if\ } o_t = y_n\\
0 & & otherwise.
\end{array}
\right.
\nonumber
\end{eqnarray}
{\bf Step 3 Parameter update and log-likelihood calculation} 
\par Update the set of parameters as $\mathrm{\Lambda} = \hat{\mathrm{\Lambda}}$ using the result of {\bf Step 2}. Calculate the probability that outputs the observation sequence $\vec{o}_{1:T}$ from the current model, and finally calculate the log-likelihood as
%
%%%%%%%%%%%%%%%%%%%%%%%
% Equation8
%%%%%%%%%%%%%%%%%%%%%%%
%
\begin{eqnarray}\label{eq:eq7}
\hat{\theta} \ \ = \ \ {\argmax_{\theta}} \log P(\vec{o}_{1:T}) & = & \log \sum_{j=1}^{M} \alpha_{T}(j,d_j),
\end{eqnarray}
where {$\alpha_{T}(j,d_j)$ is calculated {using} \Eqref{eq:eq3} when $t=T$ at the end of the sequence, and} $\hat{\theta}$ is the updated log-likelihood probability. \\
{\bf Step 4 Convergence judgement}
\par 
Judge whether the estimation process converges by evaluating that the amount of increase from 
the previous likelihood $\theta$ 
to 
the updated likelihood $\hat{\theta}$ in {\bf Step 3} is less than a predefined threshold $\epsilon$ as 
%
%%%%%%%%%%%%%%%%%%%%%%%
% Equation7'
%%%%%%%%%%%%%%%%%%%%%%%
%
\begin{eqnarray}
\label{eq:eq7_2}
\hat{\theta} - \theta & < & \epsilon.
\nonumber
\end{eqnarray}
If the condition above is satisfied, {then} the process is terminated{. O}therwise {\bf Step 2} and {\bf Step 3} are iterated until the amount of increase converges.

%
%
% 
% 
%%%%%%%%%%%%%%%%%%%%%%%%%%%%%%%%%%
% 
% IV-C.	Recognition Using HSMM
%%%%%%%%%%%%%%%%%%%%%%%%%%%%%%%%%%
\subsection{Recognition using HSMM}
\label{subsec:Recognition}
For the recognition phase that finds the model that is most likely to generate a given target observation sequence, the probability of generating {an} observation sequence plays {a fundamentally important} role. For this purpose, we first assume that a {\it label} is {assigned} appropriately into each group of sequence in advance{. T}he recognition step is defined to seek the most suitable label for a given group of sequence by calculating the label of the model that has the maximum probability as a recognition result. The probability of generating the target observation sequence is calculated using the forward algorithm used in HMM. For each model, it recursively calculates the forward variable and the probability for each state using $P(\vec{o}_{1:T})=\sum{_{i=1}^M} \alpha_T(i,d_i)$, which is the marginal probability distribution, where 
%
%%%%%%%%%%%%%%%%%%%%%%%
% Equation8
%%%%%%%%%%%%%%%%%%%%%%%
%
% Revised 20170327
\begin{eqnarray}\label{eq:eq8}
\alpha_t(j,d_j)&=&
\left[ {\sum_{i=1}^{M} \alpha_{t-{d_j}}(i,d_i)a_{(i,d_i)(j,d_j)}} \right] b_{{j,d_j}}(\vec{o}_{t-d_{{j}}+1:t}).
\end{eqnarray}
%
%%%%%%%%%%%%%%%%%%%%%%%
%

Here, we {designate} the probability {explicitly} as $P(\vec{o}_{1:T}^{*} | \mathrm{\Lambda}^z)$ using the parameter set of model $z$, i.e., $\mathrm{\Lambda}^z$, where $z \in \{1, 2, \cdots, Z\}$ and $Z$ {are} the total number of models. Finally, the label that has the maximum $P(\vec{o}_{1:T})$ for the observation sequence is selected as the recognition result. {Consequently}, the model $z^*$ that has the maximum probability $P(\vec{o}_{1:T}^{*} | \mathrm{\Lambda}^z)$ among all $Z$ models is selected {as a result} of the recognition.

% 
% 
% 
%
% 
% 
%
%%%%%%%%%%%%%%%%%%%%%%%%%%%%%%%%%%
% 
% V.		Duration and Interval in HSMM
%%%%%%%%%%%%%%%%%%%%%%%%%%%%%%%%%%
\section{State Interval Modeling in HSMM}
\label{sec:ModelingDurationAndIntervalInHSMM}
 
\begin{figure}[htb]
\begin{center}
\includegraphics[width=0.95\columnwidth]{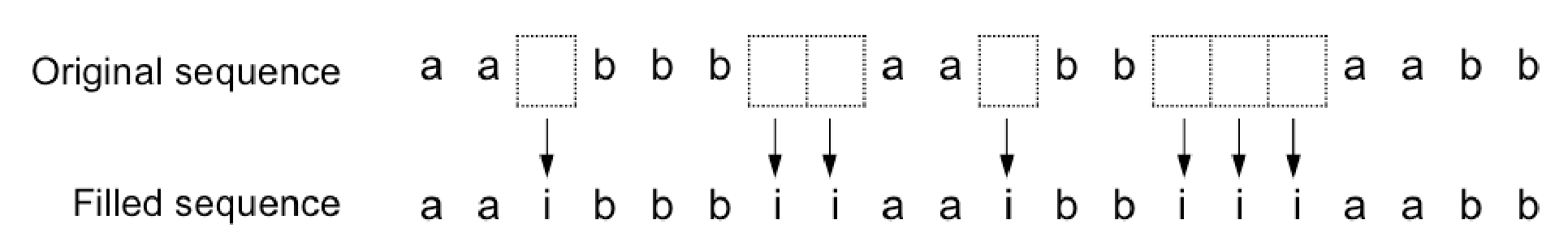}
\caption{Representation of state interval in a sequence.}
\label{fig:IntervalRepresentation}
\end{center}
\end{figure}

This section {presents investigation of} how to model {a} state interval in a model using HSMM. Before explaining the details, we describe how to represent state interval in a sequence. The baseline HSMM model ignores the period when no event is observed because the occurrence of events and the order of the events are {necessary} for sequential data modeling. However, we also {consider} this period {the} no-observation period because {it} is also {necessary} to model sequential data as described in \Secref{sec:RequirementForModel}. Therefore, we regard this period as {the} state interval in this paper, and assign a new symbol ``interval symbol" to this period. \Figref{fig:IntervalRepresentation} {portrays} an example of the state interval representation, where ``{\bf a}" and ``{\bf b}" are symbols {that are} actually observed in the original sequence, and ``{\bf i}" is the interval symbol used to fill {the} state interval. \Secref{subsec:ModelingVariation} examines the approaches for modeling state interval using HSMM{. T}he issues that arise {because of} the filled sequence with state interval are addressed in \Secref{subsec:ProblemsOfIntervalModeling}.

%%%%%%%%%%%%%%%%%%%%%%%
% Figure 3
%%%%%%%%%%%%%%%%%%%%%%%
%
%
% 
% 
%
%%%%%%%%%%%%%%%%%%%%%%%%%%%%%%%%%%
% 
% V-A.	Modeling Variation
%%%%%%%%%%%%%%%%%%%%%%%%%%%%%%%%%%
% 
\subsection{Two Approaches for State Interval Modeling}
\label{subsec:ModelingVariation}
\begin{figure}[ht]%[htbp]
\begin{center}
\includegraphics[width=0.95\columnwidth]{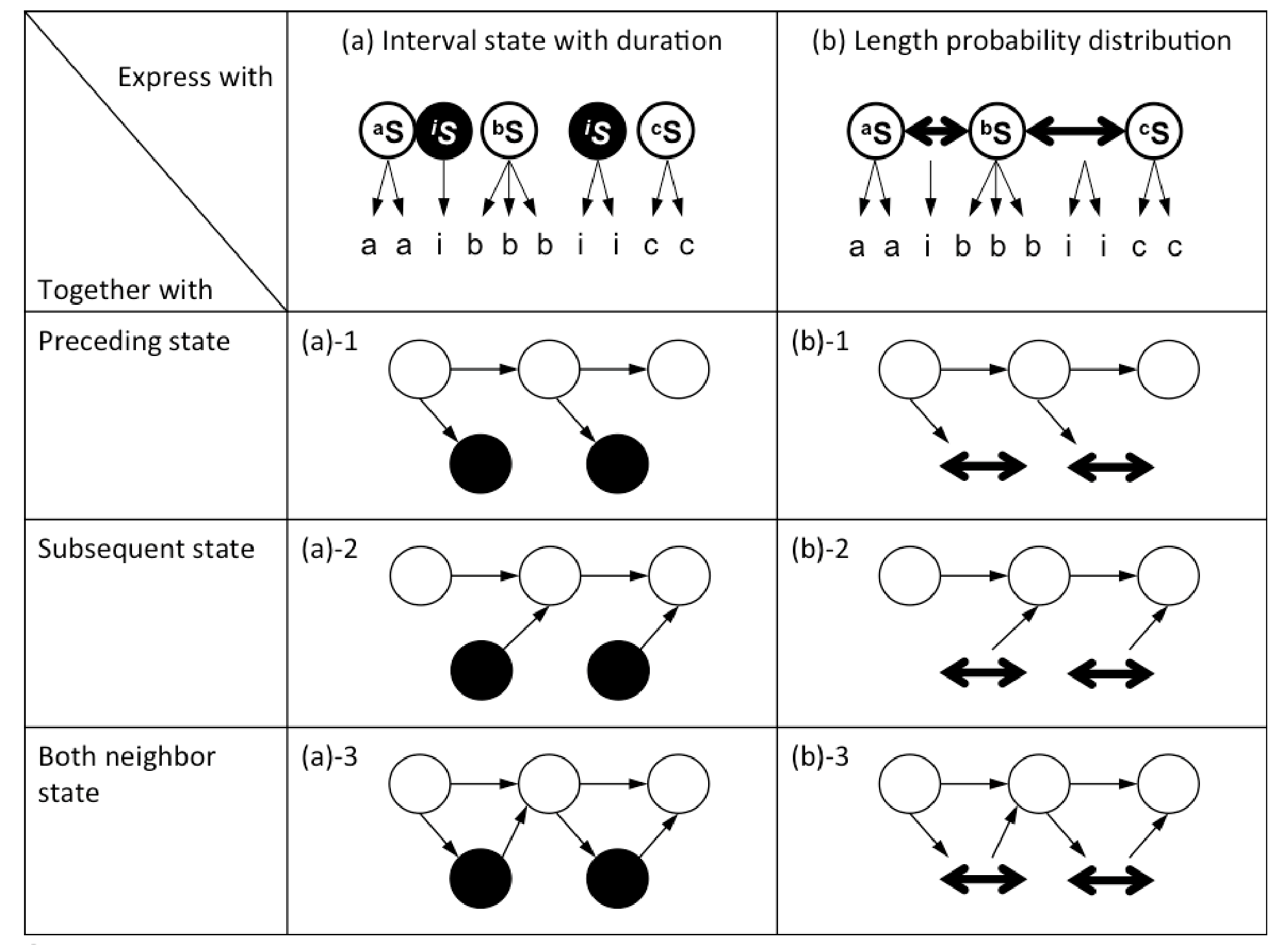}
\caption{Two approaches for {the} state interval. {The c}ircle represents a state and $\rightarrow$ represents the transition from the left state to the right state. Circle filled with black and $\leftrightarrow$ represent {the} state interval.}\label{fig:MethodsComparison}
\end{center}
\end{figure}

To treat state interval with HSMM, two approaches can be {considered} as {shown} in \Figref{fig:MethodsComparison}. One represents {the} state interval as a new state node, which is represented as a black node as \Figref{fig:MethodsComparison}(a). {E}ach state of HSMM can represent its duration {for} staying in {a} single state{. Therefore,} this new approach describes the length of {the} state interval by introducing the new state node that explicitly indicates {the} state interval. {However}, the other approach represents {the} state interval as a new probabilistic parameter as shown in \Figref{fig:MethodsComparison}(b).

For both approaches, three variations to model {the} state interval can be considered. The first approach models {the} state interval with the preceding state ((a)-1, (b)-1){;} the second models it with the subsequent state ((a)-2, (b)-2). The last variation models the length of {the} interval with both preceding and subsequent states ((a)-3, (b)-3). Compared among three variations, the first two models have connection with only one state whereas the last one ((a)-3, (b)-3) has connections with two states. Therefore, (a)-3 and (b)-3 can model the sequential data more precisely.

%
%%%%%%%%%%%%%%%%%%%%%%%
% Figure 4
%%%%%%%%%%%%%%%%%%%%%%%
%

%%%%%%%%%%%%%%%%%%%%%%%
% 
% 
% 
%
%%%%%%%%%%%%%%%%%%%%%%%%%%%%%%%%%%
% 
% V-B. Problems of Interval Modeling
%%%%%%%%%%%%%%%%%%%%%%%%%%%%%%%%%%
% 
\subsection{Problems of State Interval Modeling}
\label{subsec:ProblemsOfIntervalModeling}

\begin{figure}[ht]%[htbp]
\begin{center}
\includegraphics[width=0.9\columnwidth]{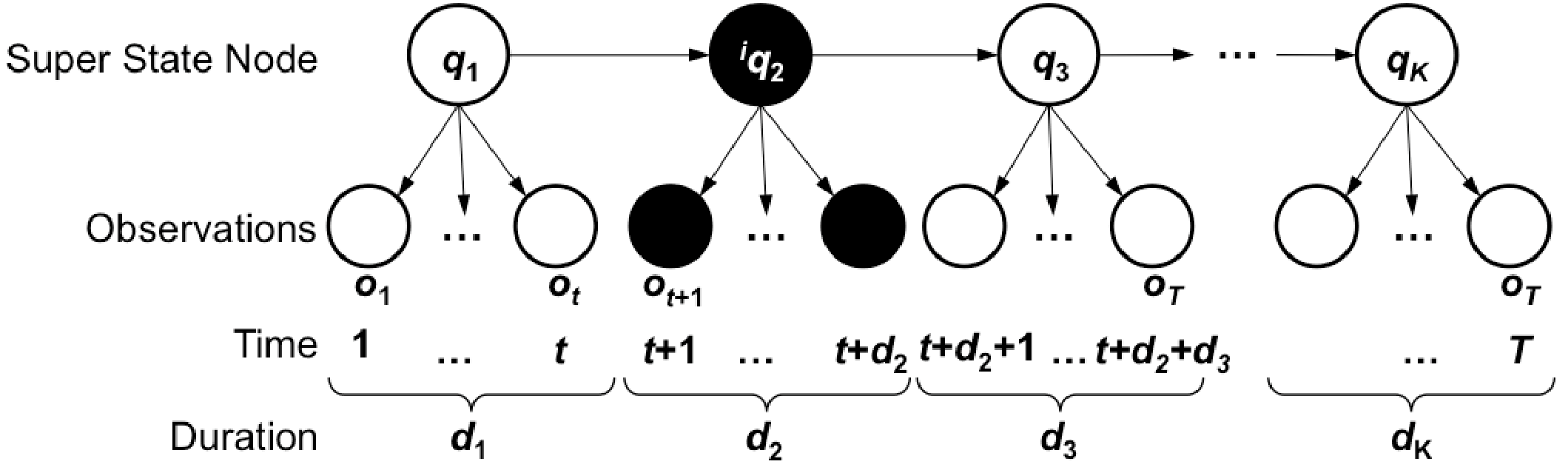}
\caption{HSMM with {an} interval state.}
\label{fig:ISHSMM}
\end{center}
\end{figure}

\begin{figure}[htbp]
\begin{center}
    \begin{subfigure}[Original sequence and transition frequency.]
           {\includegraphics[width=0.95\columnwidth]{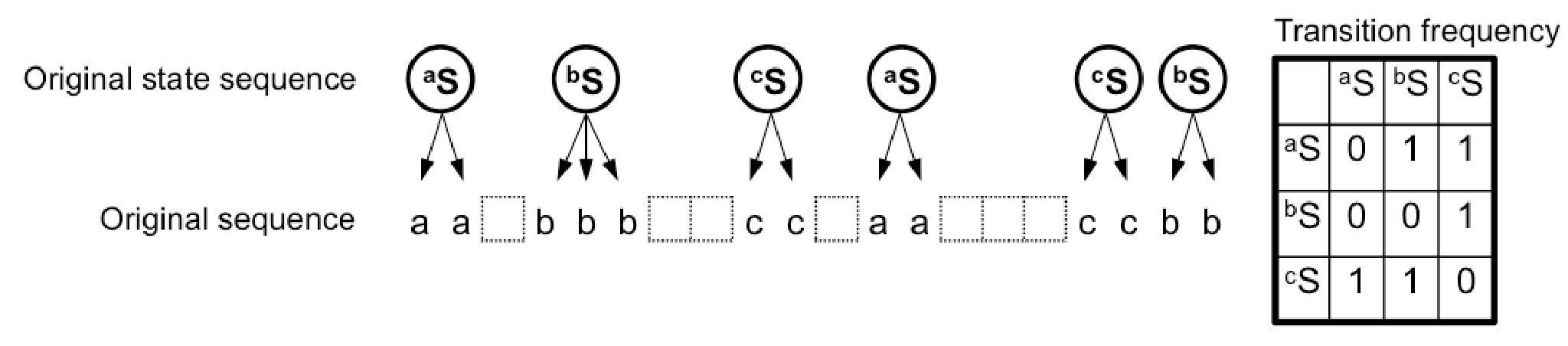}}
    \end{subfigure}
    \begin{subfigure}[Filled sequence and transition frequency.]
           {\includegraphics[width=0.95\columnwidth]{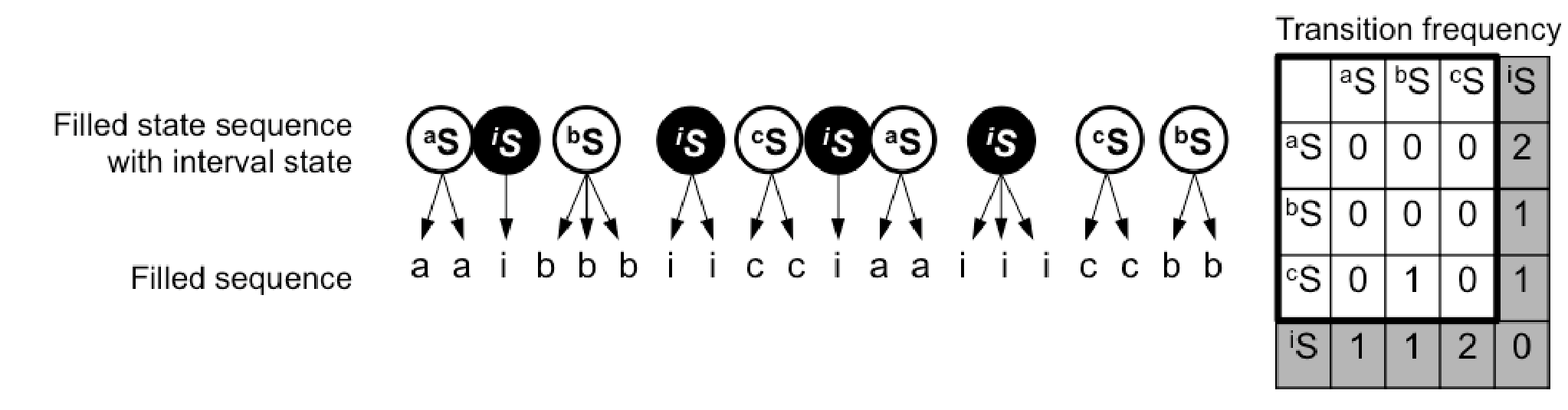}}
    \end{subfigure}
\caption{Problem of sequence with {a} state interval.}
\label{fig:ProblemOfInterval}
\end{center}
\end{figure}

Before describing the proposed models, the technical issues for the state interval modeling in each approach in the preceding subsection are explained. The structure of the first approach is {presented} in \Figref{fig:ISHSMM}, where the interval state node is presented as a black node $^iS$. Although this approach handles {the} state interval in a simple way, it causes {large} bias in the transition probability when there are many groups of 
terms {of} observed interval symbols 
in a sequence as shown in \Figref{fig:ProblemOfInterval}. 
% Details of the problem are examined using an example presented in the same figure. 
\Figref{fig:ProblemOfInterval}(a) {presents} an example sequence for the explanation. Each sequence shows the original observation sequence and the state sequence. \Figref{fig:ProblemOfInterval}(b) {presents} an example sequence filled with state interval nodes of interval symbol {\bf i}. The tables represented at the right of the figure show the transition frequency from a state to another state calculated using the original/complemented sequence. {Whereas} the states described in a vertical line in the table show the ``from" states, the states in a horizontal line show the ``to" state. The table in (a) shows the transition frequency calculated using the original state sequence{. T}he table in (b) shows the transition frequency calculated using the converted state sequence filled with interval states. Accordingly, the results reveal that the transition frequency in the {cells in the bold-framed area} except for gray painted cells falls dramatically to lower level, i.e., nearly zero. This means that, the introduction of the interval state node causes a deviation to the original transition probability{. T}he resultant new model fails to represent the transition sequence properly.

%
%%%%%%%%%%%%%%%%%%%%%%%
% Figure 5 
%%%%%%%%%%%%%%%%%%%%%%%
%
% 
%
%
%%%%%%%%%%%%%%%%%%%%%%%
%
%
%%%%%%%%%%%%%%%%%%%%%%%
% Figure 6
%%%%%%%%%%%%%%%%%%%%%%%
%

%%%%%%%%%%%%%%%%%%%%%%%
% 

{F}or the second approach in the preceding subsection, the manner of representing {a} state interval with the new probabilistic parameter ``interval length probability" must be defined. Considering {the} application data, the model is expected to be found such that sequential data have a similar sequential pattern with similar state duration and {similar} state interval. {Therefore}, it is {necessary} to model {the} state duration and {the} state interval with {representation of} the similarity of its time length. 
Therefore, the second approach defines how to represent the new parameter for state duration and how to model the parameters with {the} original HSMM in a probabilistic {manner}.

Addressing these problems, {finally, we} propose two extended models in the following sections{: an} interval state hidden semi-Markov model (IS-HSMM) as the first approach, and {an} interval length probability hidden semi-Markov model (ILP-HSMM) for the second approach.

% 
% 
% 
% 
%%%%%%%%%%%%%%%%%%%%%%%%%%%%%%%%%%
% 
% VI.	Interval State HSMM (IS-HSMM)
%%%%%%%%%%%%%%%%%%%%%%%%%%%%%%%%%%
\section{Interval State Hidden Semi-Markov Model (IS-HSMM)}
\label{sec:HSMMwithIntervalState}

\begin{figure}[htb]%[htbp]
\begin{center}
\includegraphics[width=0.95\columnwidth]{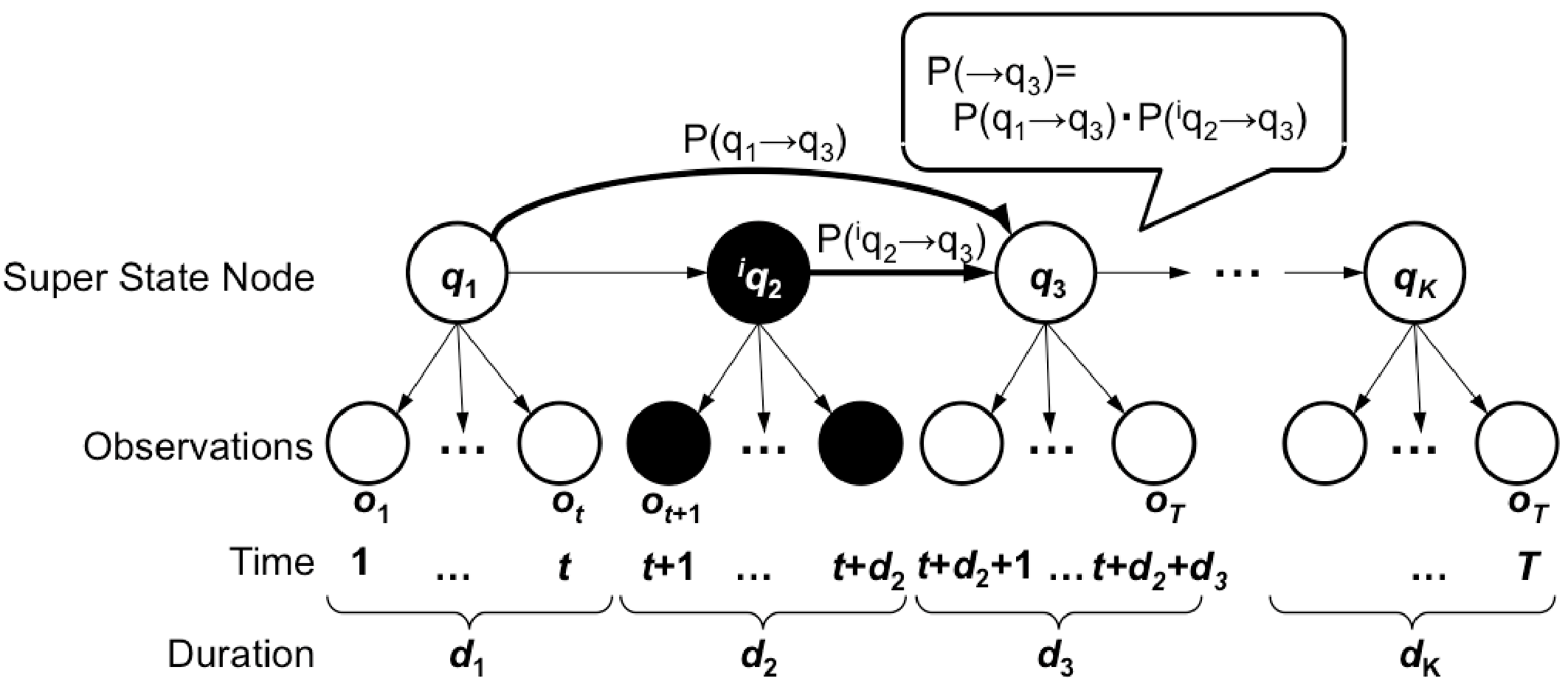}
\caption{Conceptual structure of IS-HSMM with {an} interval state node using two transition probabilities.}
\label{fig:eISHSMM}
\end{center}
\end{figure}

{Actually,} HSMM handles {the} state interval in a simple way because the interval symbol is replaced with the new interval state node as described in \Secref{sec:ModelingDurationAndIntervalInHSMM}. However, we face the {difficulty} of the degradation of the accuracy of the transition probability in cases where state intervals appear frequently in the same sequence. To {re}solve this {difficulty}, we propose an extended model, IS-HSMM, to preserve the transition probability of the original sequence. \Figref{fig:eISHSMM} {presents} a conceptual structure of IS-HSMM. For easy-to-understand explanation, we {select} the first 
three states shown in \Figref{fig:eISHSMM} as an example when $q_1$ and $q_{{3}}$ are original hidden states and ${^i}q_{{2}}$ is the interval state node. {Whereas} the original HSMM infers the transition probability in the order of $q_1$, $^iq_2$, and $q_3$, the proposed IS-HSMM infers the transition probability {as} $q_3$ using two transition probabilities not only from $^iq_2$ to $q_3${,} but also from the previous $q_1$ to $q_3$ to preserve the transition of the original sequence. This is {a} noteworthy feature of IS-HSMM. This section {explains} how to train and how to recognize the model as follows.

%%%%%%%%%%%%%%%%%%%%%%%
% 
%
% 
% 
% 
%
%%%%%%%%%%%%%%%%%%%%%%%
% Figure 7
%%%%%%%%%%%%%%%%%%%%%%%
%
%%%%%%%%%%%%%%%%%%%%%%%
% 

%
%%%%%%%%%%%%%%%%%%%%%%%%%%%%%%%%%%
% 
% VI-A.	Training
%%%%%%%%%%%%%%%%%%%%%%%%%%%%%%%%%%
% 
\subsection{Model Training in IS-HSMM}
\label{subsec:TrainingOfISHSMM}
The difference against the baseline HSMM model appears in the calculation of the forward variables and backward variables in the recursive calculation step. The state transition probability from state $i$ to state $j${,} where the interval state $^is$ is inserted between state $i$ and state $j${,} is defined as
%
%%%%%%%%%%%%%%%%%%%%%%%
% Equation9
%%%%%%%%%%%%%%%%%%%%%%%
%
\noindent
%
%%%%%%%%%%%%%%%%%%%%%%%
% Equation9
%%%%%%%%%%%%%%%%%%%%%%%
%
% Revise 20170326
\begin{eqnarray}\label{eq:eq9}
a_{(i,d_i)(^is,^id)(j,d_j)}
&:=& P(S_{t+{^id+}1:t+{^id+}d_j}=j | S_{t+1:t+^id}=^is ,~ S_{t-d_i+1:t}=i)\nonumber\\
&:=& P(S_{t+1:t+d_j}=j | S_{t-^id+1:t}=^is ,~ S_{t-d_i-d_i+1:t-^id}=i)\nonumber
\end{eqnarray}
where the duration of interval state $^is$ is denoted as $^id$($>0$).
%{ ($>0$), assuming that the two transitions are independent.} 
{The respective} durations of state $i$ and $j$ are $d_i$ and $d_j$. 
%{Because the transition from $i$ to $^is$ is assumed to be treated when the state is $^is$, }
{The transition $a_{(i,d_i)(^is,^id)(j,d_j)}$ is calculated with the transition from $^is$ and the preceding state $s_i$ only when calculating after $^is$. Therefore, }
%
%%%%%%%%%%%%%%%%%%%%%%%
%
the forward variable{, where the current state is $j$ and the preceding state is $^is$,} is calculated {using the further preceding state $i$ based on the second-order HMM \cite{He1988}} as
%
%%%%%%%%%%%%%%%%%%%%%%%
% Equation10
%%%%%%%%%%%%%%%%%%%%%%%
%
\noindent
%
%%%%%%%%%%%%%%%%%%%
% equation
%%%%%%%%%%%%%%%%%%%
% 
\begin{eqnarray}\label{eq:eq10}
\alpha_t((^is,^id),(j,d_j)) &=&
\sum_{i \in \{S\} \setminus \{j,^is\}} \sum_{d_i \in D} \alpha_{t-^id}((i,d_i),(^is,^id))  
\cdot a_{(i,d_i)(^is,^id)(j,d_j)} b_{j,d_j}(o_{t-^id+1:t}) ,
\end{eqnarray}
{where $a_{(i,d_i)(^is,^id)(j,d_j)}$ is updated the following equation.}
%% add this equation
%
%%%%%%%%%%%%%%%%%%%
% equation
%%%%%%%%%%%%%%%%%%%
% 
\begin{eqnarray}\label{eq:updateSecondOrderTransition}
a'_{(i,d_i)(^is,^id)(j,d_j)}&=&
\sum_{t=1}^{T-d_j-^id} \alpha_{t+^id}((i,d_i),(^is,^id))  a_{(i,d_i)(^is,^id)(j,d_j)} 
\cdot b_{j,d_j}(o_{t+^id+d_j}) \beta_{t+^id+d_j}(j,d_j)  \nonumber \\
&&\big/ \sum_{t=1}^{T-d_j-^id} \sum_{j \in \{S\} \setminus \{i,^is\} } \alpha_{t+^id}((i,d_i),(^is,^id)) a_{(i,d_i)(^is,^id)(j,d_j)}
\cdot b_{j,d_j}(o_{t+^id+d_j}) \beta_{t+^id+d_j}(j,d_j) . \nonumber
\end{eqnarray}
% 
%
%%%%%%%%%%%%%%%%%%%%%%%
%
% {when the preceding state is $^is$.} 
Then, the backward variable {where the preceding state is $^is$ is calculated as general first-order transition probabilities} 
expressed as
%
%%%%%%%%%%%%%%%%%%%%%%%
% Equation11
%%%%%%%%%%%%%%%%%%%%%%%
%
\noindent
% Revise 20170330
\begin{eqnarray}\label{eq:eq11}
\beta_t(j,d_j) &=&
\sum_{i \in \{S\} \setminus \{j\}} \sum_{^id \in D}
a_{(j,d_j)(^is,^id)} b_{^is,^id}(\vec{o}_{t+1:t+^id}) \beta_{t+^id}(^is,^id).
\end{eqnarray}
%
%%%%%%%%%%%%%%%%%%%%%%%
%
Finally, the transition probability and the emission probability 
are updated using \Eqref{eq:eq10} and \Eqref{eq:eq11} by {calculating the state transition probability using \Eqref{eq:eq9} and} assigning the forward and backward variables obtained {respectively using} \Eqref{eq:eq5} and \Eqref{eq:eq6}.

%
% 
%
%%%%%%%%%%%%%%%%%%%%%%%%%%%%%%%%%%
% 
% VI-B.	Recognition
%%%%%%%%%%%%%%%%%%%%%%%%%%%%%%%%%%
% 
\subsection{Recognition using IS-HSMM}
\label{subsec:RecognitionOfISHSMM}
{Although} calculation of the probability follows the original HSMM when the preceding state is not the interval state node, it differs when the preceding state is the interval state node. The probability of the observation sequence when the preceding state is the interval state node is calculated as
%
%%%%%%%%%%%%%%%%%%%%%%%
% Equation12
%%%%%%%%%%%%%%%%%%%%%%%
%
$P(\vec{o}_{1:T})=\sum_{i=1}^{M} \alpha_T(i,d_i)$,
where {\Eqref{eq:eq8} and the follows:}
% Revised 20170327
\begin{eqnarray}\label{eq:eq12}
\alpha_t(j,d_j) =
\left[ \sum_{i=1}^{{M}}
\alpha_{t-{d_j-^id}}(i,d_i)a_{(i,d_i)(^is,^id)} {\cdot a_{(^is,^id)(j,d_j)}} \right] 
b_{^is,^id}(\vec{o}_{t-^id-d_j+1:t-d_j}) \cdot b_{j,d_j}(\vec{o}_{t-d_j+1:t}),\ \ 
\end{eqnarray}
%
%%%%%%%%%%%%%%%%%%%%%%%
%
{where the preceding state is $^is$}. The overall algorithm is presented in \Algref{alg:alg2}.

%%%%%%%%%%%%%%%%%%%%%%%
%
%
%%%%%%%%%%%%%%%%%%%%%%%
% pseudocode (IS-HSMM proposed)
%%%%%%%%%%%%%%%%%%%%%%%
\begin{algorithm}[htb]%[htb][H]
\caption{Algorithm for training and recognition in IS-HSMM.}     
\label{alg:alg2} 
\begin{algorithmic}[1]
\REQUIRE {Input}\\
Training sequences: $\vec{o}_{1:T_r}^{z}=\{o_1^z, \cdots, o_{T_r}^z\}$, \\
Testing sequences: $\vec{o}_{1:T_t}^{*}=\{o_1^{*}, \cdots, o_{T_t}^{*}\} $. \\
{($Z$ is the number of training sequences.)}\\
{($H$ is the number of recursive calculation.)}

\ENSURE {\bf Training phase}
\FOR {$z=1$ to $Z$} 
\STATE{
Assign random values to the HSMM parameters $\mathrm{\Lambda}=\{\mat{A}, \mat{B}, \mat{$\pi$}\}$, and $\alpha_{t(j,d_j)}$ and $\beta_{t(j,d_j)}$.
\\ 
}
\FOR {$h=1$ to $H$} 
\FOR {$t=1$ to $T_r$} 
\IF{$o_{t-1}$ is interval symbol}
\STATE{
 Calculate $\alpha_{t(j,d_j)}$ and $\beta_{t(j,d_j)}$ with joint probability from $i$ and $^is$ using \Eqref{eq:eq10} and \Eqref{eq:eq11}.
}
\ELSE
\STATE{
 Calculate $\alpha_{t(j,d)}$ and $\beta_{t(j,d_j)}$ with preceding state $i$ using \Eqref{eq:eq3} and \Eqref{eq:eq4}.
}
\ENDIF
\STATE{
 Update parameters $\mathrm{\Lambda}$.
}
\ENDFOR \\
\STATE{
 Calculate $\theta_h$ using \Eqref{eq:eq7} with \Eqref{eq:eq10}.
 \IF{$\theta_h - \theta_{h-1} < \epsilon$}
 \STATE{{\bf break}}
 \ENDIF
}
\ENDFOR \\
\ENDFOR \\
\ENSURE {\bf Testing phase}
\FOR {$z=1$ to $Z$}
\FOR {$t=1$ to $T_t$} 
\IF{$o_{t-1}$ is {the} interval symbol}
\STATE{
 $\mathrm{\Lambda}^z$ $\leftarrow$ parameter $\mathrm{\Lambda}$ of model $z$ 
 with joint probability from $j$ and $^is$.
}
\ELSE
\STATE{
 $\mathrm{\Lambda}^z$ $\leftarrow$ parameter $\mathrm{\Lambda}$ of model $z$ with preceding state $j$.
}
\ENDIF
\STATE{
  Calculate $\alpha_t(j,d_j)$ using \Eqref{eq:eq8} with \Eqref{eq:eq12}.
}
\ENDFOR
\STATE{
  Calculate $P(o_{1:T_t} | \mathrm{\Lambda}^z)$ using $\alpha_t(j,d_j)$.
}
\ENDFOR
\STATE{
 Select the model $z^*$ that has the maximum value for $P(\vec{o}_{1:T_t}^{*} | \mathrm{\Lambda}^z)$.
}
\STATE{
 {\bf Return} Model $z^*$ and its probability $P(\vec{o}_{1:T_t}^{*} | \mathrm{\Lambda}^{z^*})$.
}
\end{algorithmic}
\end{algorithm}
%%%%%%%%%%%%%%%%%%%%%%%
%
%%%%%%%%%%%%%%%%%%%%%%%%%%%%%%%%%%
% 
% VII.	Duration and Interval Hidden Markov Model ({ILP-HSMM})
%%%%%%%%%%%%%%%%%%%%%%%%%%%%%%%%%%
% 
\section{Interval Length Probability HSMM (ILP-HSMM)}
\label{sec:ILPHSMM}

\begin{figure}[tb]%[htbp]
\begin{center}
\includegraphics[width=0.95\columnwidth]{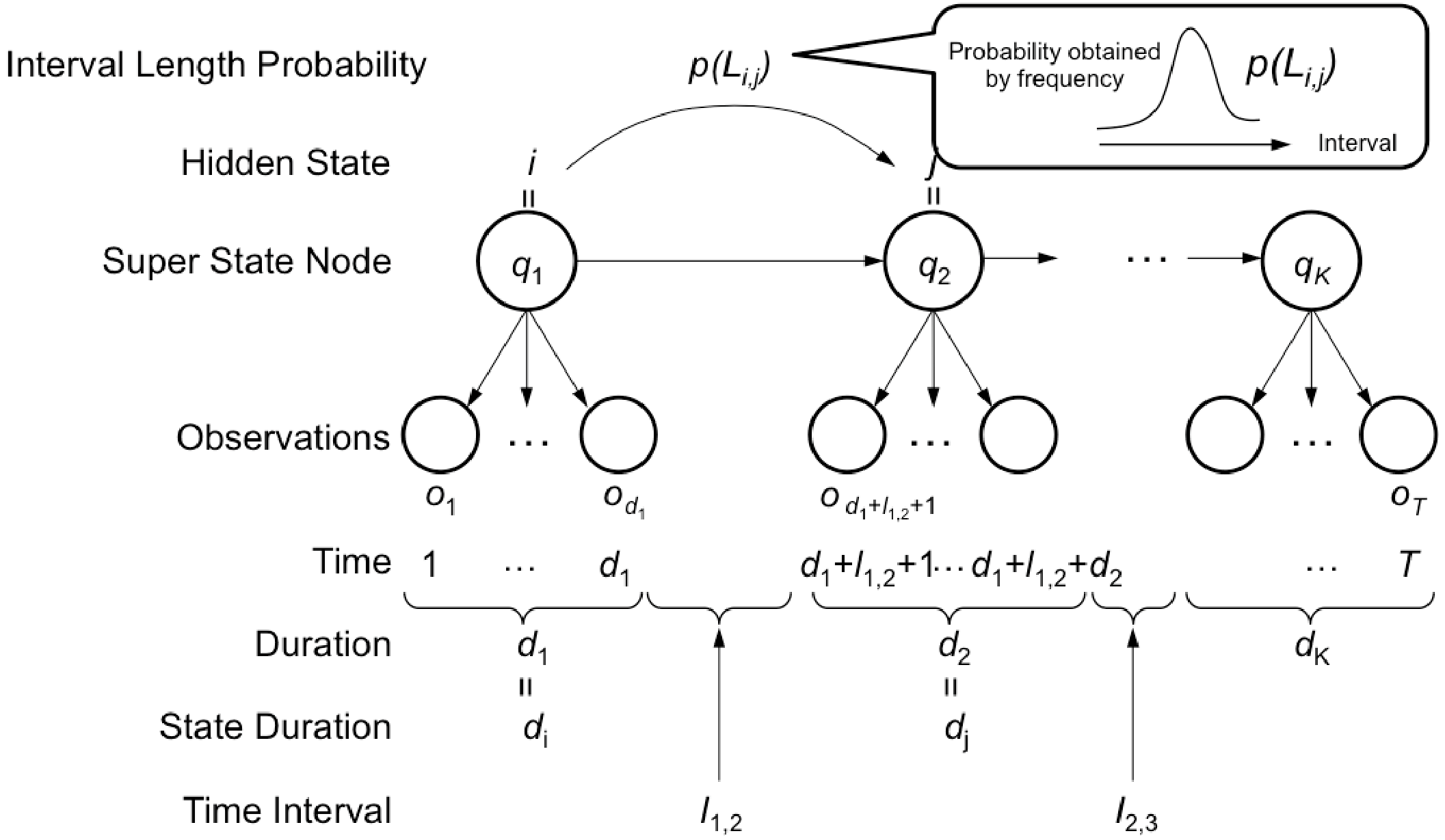}
\caption{Conceptual structure of ILP-HSMM using state interval probability. 
}
\label{fig:ILPHSMM}
\end{center}
\end{figure}

This section {presents} ILP-HSMM, {which} newly introduces interval length probability to the transition probability to handle {the} state interval between two states. {It is noteworthy} that the interval length probability corresponds to the probability density distribution of interval length of two states, to be technically precise. The distinct difference between HSMM and ILP-HSMM is that, whereas state $j$ starts {immediately} after the end time of state $i$ in the original HSMM, state $j$ starts after a length of time, $L_{i,j}$, passes since the end time of state $i$ in ILP-HSMM. The conceptual model structure of ILP-HSMM is {presented} in \Figref{fig:ILPHSMM}. Although the ILP-HSMM {structure} is similar to that of HSMM {presented} in \Figref{fig:HSMM}, the interval length probability is newly added to HSMM as shown in \Figref{fig:ILPHSMM}, where $L_{i,j}$ represents the time difference between the end time of state $i$ and the beginning time of state $j$. {It is noteworthy} that the total time length of the observation sequence $T$ varies because of its dependency on the length of state duration and interval, leading to $T=\Sigma^{K}_{k=1}(d_k+l_{k-1,k})$, where $l_{k-1,k}$ 
is the time difference between the end of $q_{k-1}$ and the beginning of $q_{k}$. The subsequence section {presents a description of} how to model and how to recognize given datasets using ILP-HSMM.
% 
% 
%%%%%%%%%%%%%%%%%%%%%%%
% Figure 8
%%%%%%%%%%%%%%%%%%%%%%%
% 
% 
% 
%%%%%%%%%%%%%%%%%%%%%%%
% 
% VII-A. Training Sequential Data Model
%%%%%%%%%%%%%%%%%%%%%%%
\subsection{Model Training (Inference) in ILP-HSMM}
\label{sec:ILPHSMMTraining}
\begin{figure}[tb]%[htbp]
\begin{center}
\includegraphics[width=0.95\columnwidth]{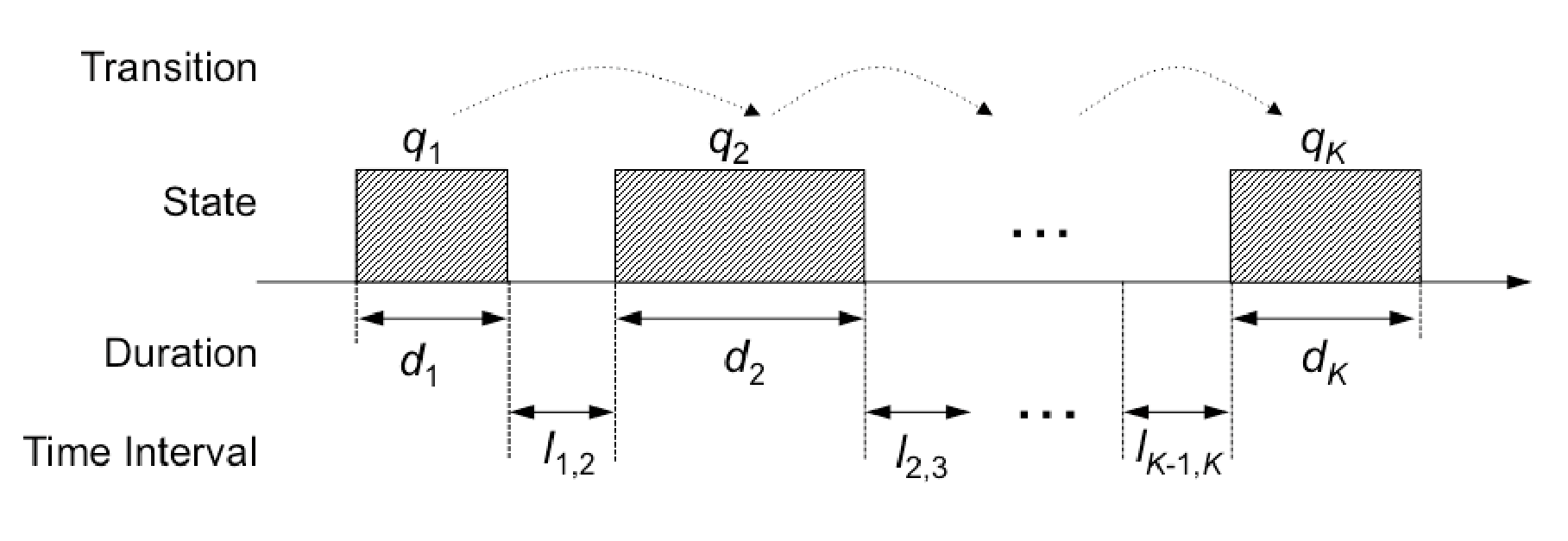}
\caption{Sequential data and representations.}
\label{fig:SequentialData}
\end{center}
\end{figure}

\Figref{fig:SequentialData} {presents} example data and representations used {hereinafter} for explanation. The slash line patterned blocks represent the data sequence of {the} training dataset. First, the probability density distribution of the interval length of $L_{i,j}$ is expressed by the Gaussian distribution $p(L_{i,j})$ as
%
%%%%%%%%%%%%%%%%%%%%%%%
% Equation (14)
%%%%%%%%%%%%%%%%%%%%%%%
%
\begin{eqnarray}\label{eq:eq13}
p(L_{i,j})~=~\frac{1}{\sqrt{2\pi\sigma^2}}e^{- { \frac{(L_{i,j}- \mu)^2}{2\sigma^2} }},
\end{eqnarray}
where $\sigma$ and $\mu$ {respectively} present the variance and the mean of $L_{i,j}$. {It is noteworthy} that the Gaussian distribution is adopted as the probability density distribution{,} for simplicity. However, other distributions and functions {were} adopted for ILP-HSMM without changing any other parameter. Accordingly, the set of parameters used in ILP-HSMM is defined as
%
%%%%%%%%%%%%%%%%%%%%%%%
% Equation (15)
%%%%%%%%%%%%%%%%%%%%%%%
%
\begin{eqnarray}\label{eq:eq14}
\mathrm{\Lambda} := \{\mat{A}, \mat{B}, \vec{\pi}, \mat{L}\},
\nonumber
\end{eqnarray}
where the elements of the parameter $\mathrm{\Lambda}$ take on $\mat{A}(i,j) = a_{(i,d_i)(j, d_j)}$, $\mat{B}(j,n) = b_{(j, d_j)}(\vec{o}_{1: d_j})$, and $\vec{\pi}(i) = \pi_{j, d_j}$, where $d_i \in D$ represents the duration of state $i$ 
described in \Secref{subsec:Notation}. Furthermore, $\mat{L} \in \mathbb{R}^{M \times M}$ is the matrix that consists of the interval length probabilities, i.e., the probability density distributions of {the length of state interval}, where $\mat{L}(i,j) = p(L_{i,j})$. The transition and emission probabilities are defined as the same as those in HSMM. The difference between HSMM and ILP-HSMM is to consider the parameter of $p(L_{i,j})$.

The range of $L_{i,j}$ in \Eqref{eq:eq13} might influence either memory consumption {or} computational complexity to generate the model. There might be no $L_{i,j}$ value suitable for the observation values because of the range limitation of $L_{i,j}$ if $p(L_{i,j})$ is generated in a training period. However, if the parameter $p(L_{i,j})$ is generated every time an observation is fed to the algorithm, {then} the calculation cost can be much higher. Our motivation to introduce the interval length probability to HSMM is, as explained earlier, to find the similar part of sequential data with respect to {the} state interval and also to discriminate between the target part and the similar part. Therefore, even if the probability of $L_{i,j}$ is presumed to be zero around the skirts of the distribution, no {critically important difficulty} arises. Consequently, we introduce the boundary of the probability value $\delta_{pt}$ to {ascertain} the edge of the skirt of $p(L_{i,j})$. On generating the $p(L_{i,j})$, the calculation is terminated when the probability value becomes less than 
$\delta_{pt}$. The probability of $p(L_{i,j})$ is zero outside of the range of $\delta_{pt}$.
% 
%%%%%%%%%%%%%%%%%%%%%%%
% Figure 9
%%%%%%%%%%%%%%%%%%%%%%%
% 
%%%%%%%%%%%%%%%%%%%%%%%
% 
% 
% 
%%%%%%%%%%%%%%%%%%%%%%%%%%%%%%%%%%
% 
% VII-B. Probability Estimation for Recognition
%%%%%%%%%%%%%%%%%%%%%%%%%%%%%%%%%%
\subsection{Recognition using ILP-HSMM}
\label{sec:ILPHSMMRecognition}
The Viterbi algorithm is used to estimate the probability of a model \cite{MurphyThesis2002}. The pair of the model with the interval length probability and its label {that is} expected to be estimated are stored as candidates for estimation. The recognition label that {denotes} the estimated result is selected when the model has the maximum likelihood estimate by calculating it for each state in each model.

First, we calculate $p(L_{i,j})$ beforehand. If $L_{i,j}$ is out of the range, {then} the probability density distribution is determined as
%
%%%%%%%%%%%%%%%%%%%%%%%
% Equation (16)
%%%%%%%%%%%%%%%%%%%%%%%
%
\begin{eqnarray}\label{eq:eq15}
p(L_{i,j}) & = & \min_{i, j \in S} p(L_{i,j}) \times c,
\nonumber
\end{eqnarray}
%%%%%%%%%%%%%%%%%%%%%%%
%
where $c$ is $0 \leq c \leq 1$. Then, the forward variable for estimating the maximum likelihood is calculated as
%
%
%%%%%%%%%%%%%%%%%%%%%%%
% Equation (17) %% revise 
%%%%%%%%%%%%%%%%%%%%%%%
%
\begin{eqnarray}\label{eq:eq16}
\alpha_t(j,d_j)&=& 
\left[ \sum_{i=1}^{M} \alpha_{t-{d_j}}(i,d_i)a_{(i,d_i)(j,d_j)} \cdot p(L_{i,j}) \right]
b_{j,d_j}(\vec{o}_{t-d_{j}+1:t}).
\end{eqnarray}
%%%%%%%%%%%%%%%%%%%%%%%
%
%
%
%
%
The interval length probability is calculated simultaneously {with calculation of} the parameter of the likelihood using the transition probability recursively.

The difference between HSMM and ILP-HSMM is the capability of handling the length of the state interval between states as explained earlier. The interval length probability in ILP-HSMM {can} be integrated by introducing each interval into two pair of states to calculate the likelihood. This calculation might {produce an} additional calculation cost. Therefore, it is necessary to evaluate the calculation cost. In addition, the emission probability $b_{j, d_j}({\bf o}_{1: d_j})$ can be parametric or non-parametric. In this proposal, the relation {between the} state duration and {the} state interval is not represented in a model. For this reason, $b_{j, d_j}({\bf o}_{1: d_j})$ is handled as non-parametric, discrete, and independent of the duration. Then, $p(L_{i,j})$ is also discrete and independent of the duration and the transition probability.
%
%
%%%%%%%%%%%%%%%%%%%%%%%
% pseudocode ({ILP-HSMM} proposed) alg3:ILPHSMM
%%%%%%%%%%%%%%%%%%%%%%%
\begin{algorithm}[htb]
\caption{Algorithm for training and recognition in ILP-HSMM.}     
\label{alg:ILPHSMM} 
\begin{algorithmic}[1]
\REQUIRE {Input}\\
Training sequences: $\vec{o}_{1:T_r}^{z}=\{o_1^z, \cdots, o_{T_r}^z\}$,\\
Testing sequences: $\vec{o}_{1:T_t}^{*}=\{o_1^{*}, \cdots, o_{T_t}^{*}\} $. \\
{($Z$ is the number of training sequences.)}\\
{($H$ is the number of recursive calculation.)}

\ENSURE {\bf Training phase}
\FOR {$z=1$ to $Z$} 
\STATE{
Assign random values to the HSMM parameters $\mathrm{\Lambda}=\{\mat{A}, \mat{B}, \mat{$\pi$}, \mat{L}\}$, and $\alpha_{t(j,d_j)}$ and $\beta_{t(j,d_j)}$. \\
Initialize $p(L_{i,j})$ as $L_{i,j}=1$.\\
}
\FOR {$h=1$ to $H$} 
\FOR {$t=1$ to $T_r$} 
\STATE{
 Calculate $\alpha_{t(j,d_j)}$ and $\beta_{t(j,d_j)}$ using \Eqref{eq:eq3} and \Eqref{eq:eq4}.
}
\STATE{
 Calculate $p(L_{i,j})$ with $i$ and $j$ using \Eqref{eq:eq13}.
}
\STATE{
 Update parameters $\mathrm{\Lambda}$.
}
\ENDFOR
\STATE{
 Calculate $\theta_h$ using \Eqref{eq:eq12}.
 \IF {$\theta_h - \theta_{h-1} < \epsilon$}
 \STATE{{\bf break}}
 \ENDIF
}
\ENDFOR
\ENDFOR
\ENSURE {\bf Testing phase}
\FOR {$z=1$ to $Z$}
\FOR {$t=1$ to $T_t$} 
\STATE{
 $\mathrm{\Lambda}^z$ $\leftarrow$ parameter $\mathrm{\Lambda}$ of model $z$.
}
\STATE{
 $p(l)$ $\leftarrow$ $p(L_{i,j})$ using $\mathrm{\Lambda}^z$ with observed interval $l$.
}
\STATE{
  Calculate $\alpha_t(j,d_j)$
  using \Eqref{eq:eq8} with \Eqref{eq:eq16}. 
}
\ENDFOR
\STATE{
    Calculate $P(o_{1:T_t} | \mathrm{\Lambda}^z)$ using $\alpha_t(j,d_j)$.
}
\ENDFOR
\STATE{
Select model $z^*$ {with} the maximum value for $P(\vec{o}_{1:{T_t}}^{*} | \mathrm{\Lambda}^z)$.
}
\STATE{
{\bf Return} Model $z^*$ and its probability $P(\vec{o}_{1:T}^{*} | \mathrm{\Lambda}^{z^*})$.
}
\end{algorithmic}
\end{algorithm}
%%%%%%%%%%%%%%%%%%%%%%%
%
% 
% 
% 

%%%%%%%%%%%%%%%%%%%%%%%%%%%%%%%%%%
% 
% VIII.	Evaluations
%%%%%%%%%%%%%%%%%%%%%%%%%%%%%%%%%%
\section{Evaluations}
\label{sec:Evaluation}
This section presents a description of the performance evaluation of models. After explaining the specification{s} of the experimental data in \Secref{subsec:ExperimentData}, \Secref{subsec:ExecutionTimeEvaluation} and \Secref{subsec:RecognitionPerformanceEvaluation} {present} the {experimentally obtained results} of the execution time and recognition performance comparison among HSMM, IS-HSMM, and ILP-HSMM. Finally, we evaluate {a} reproducibility comparison between IS-HSMM and ILP-HSMM in terms of the modeling performance in \Secref{subsec:EffectivenessOfProposals}.
%
%
%%%%%%%%%%%%%%%%%%%%%%%%%%%%%%%%%%
% 
% VIII-A.	Experimental Data
%%%%%%%%%%%%%%%%%%%%%%%%%%%%%%%%%%
\subsection{Experimental Data}
\label{subsec:ExperimentData}
\begin{figure}[htb]%[htb]%[htbp]
\begin{center}
\includegraphics[width=\columnwidth]{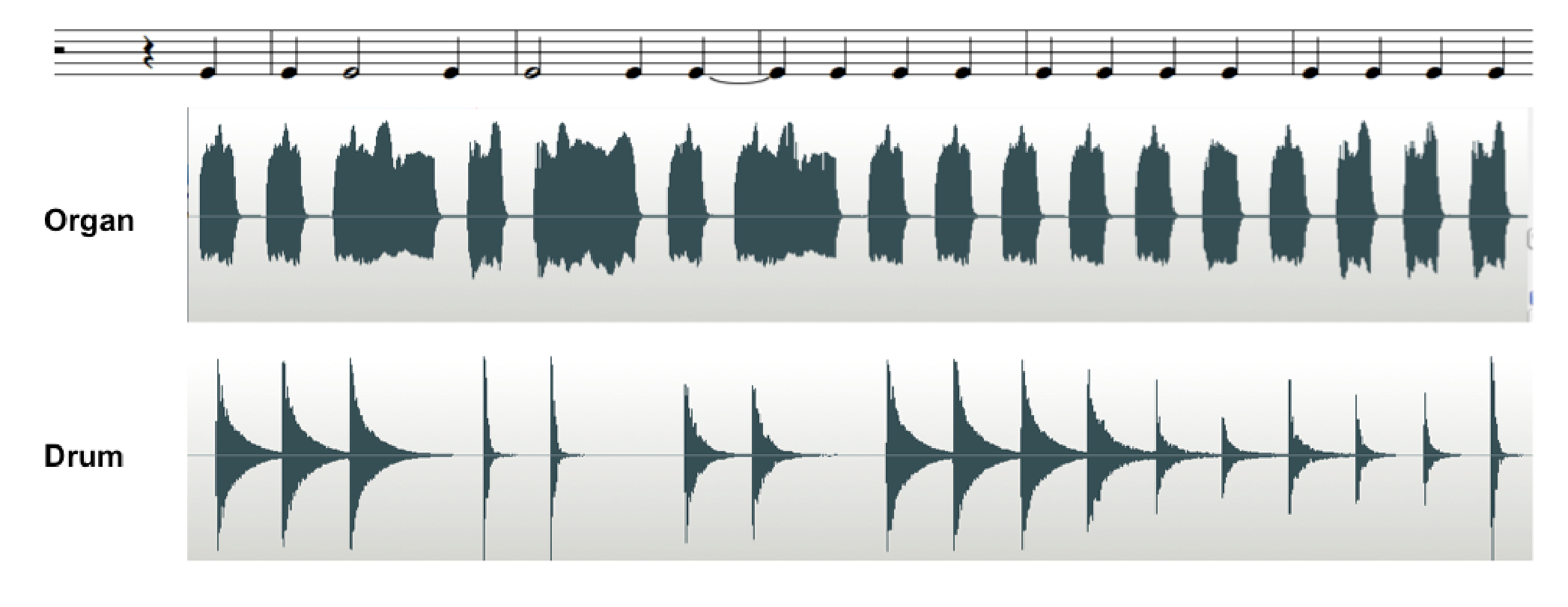}
\caption{Music sound data.}
\label{fig:MusicSoundData}
\end{center}
\end{figure}

Addressing that the sequential data contain {the} state duration and {the} state interval, we use music sound data played by instruments of different kinds. When the same music is played by the different instruments, even if the music rhythm is the same, the length of each sound for the same note differs. For example, the sound power spectrum played by {an} organ and {a} drum for the same music sound data is shown in \Figref{fig:MusicSoundData}. The horizontal axis {shows the} time{. T}he vertical axis {shows} the sound power, i.e., sound volume. {Whereas} the power of each note played by the organ is almost {identical} during the sound {resonation}, the one played by the drum decreases rapidly after tapping. We generate the observation sequence from the music sound data. The generation step is described below using the features of sound continuous time.\\
{\bf Step 1}
\par Set {thresholds} $b_1$ and $b_2$ to classify the observation symbols into three types by the level of the volume. $b_1$ is a threshold for determining whether the sound {is} ``on" or ``off", and $b_2$ is the one for classifying the power of the sound {as} ``high" {or} ``low". $(b_2 \geq b_1)$\\
{\bf Step 2}
\par For the sound power $v$ of each time, the observation sequence is generated as follows.
\par If $v \geq b_2$, then the observation symbol is ``high".\\
\par
\begin{figure}[htb]%[tb]%[htbp]
\begin{center}
\includegraphics[width=\columnwidth]{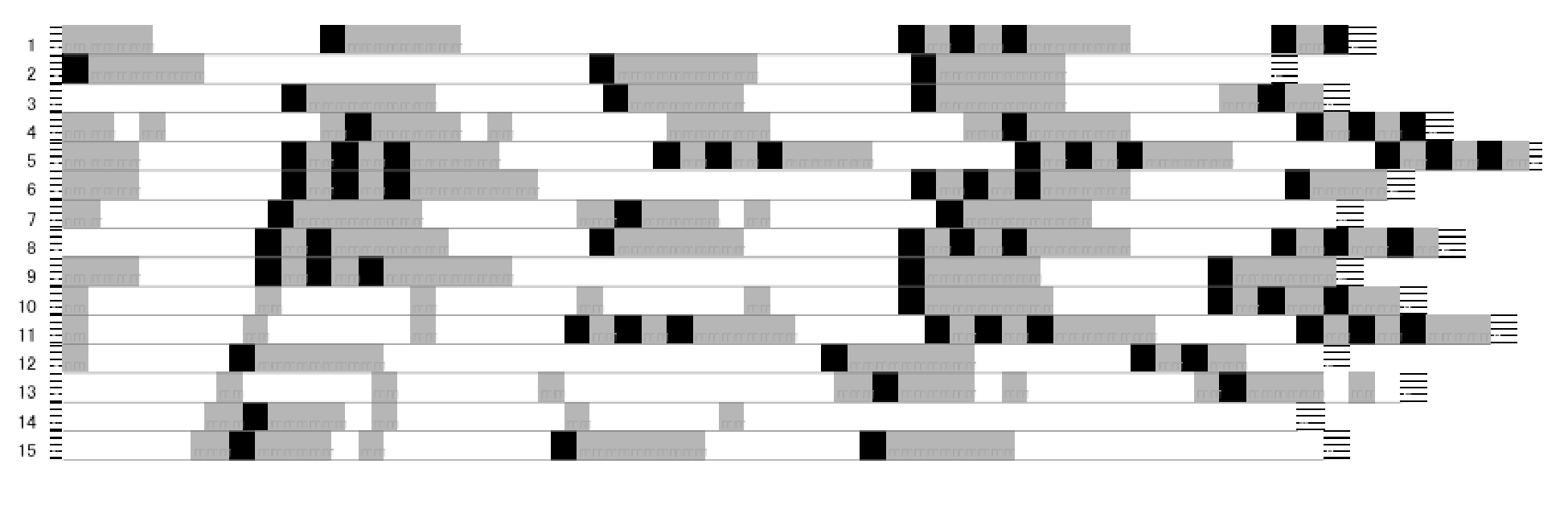}
\caption{Example sequences generated using music sound data.}
\label{fig:ExampleSequence}
\end{center}
\end{figure}

 If $b_2 > v \geq b_1$, then the observation symbol is ``low".\\
An example of observation sequence generated by the procedure described above is shown in \Figref{fig:ExampleSequence}. The black cell represents the ``high" symbol{. T}he gray cell represents a ``low" symbol{. The white-painted cells represent} the ``interval." 
To {denote} the segment of a sequence, we add ``start" and ``end" symbols to each edge of the sequence. These symbols are useful for modeling the transition from the initial state from sequences precisely. The dataset consists of 27 segmented data, which are divided into bars of the music sequence. A label is assigned for each 27 segmented data{. T}herefore the number of labels is also 27. The kinds of the instruments are a grand piano, horn, drums, acoustic guitar, flute, and pipe organ. We use the music sound data played by the instrument{s} of {the} first three kinds for training data, {whereas} the latter three kinds are used for recognition data. The number{s} of the sequential data {are} 81 for both training and recognition.

%
%%%%%%%%%%%%%%%%%%%%%%%
% Figure 10
%%%%%%%%%%%%%%%%%%%%%%%
%
%%%%%%%%%%%%%%%%%%%%%%%
%
%
%%%%%%%%%%%%%%%%%%%%%%%
% Figure 11
%%%%%%%%%%%%%%%%%%%%%%%
%
%%%%%%%%%%%%%%%%%%%%%%%
% 
% 
%%%%%%%%%%%%%%%%%%%%%%%%%%%%%%%%%%
% 
% VIII-B.	Execution Time Evaluation
%%%%%%%%%%%%%%%%%%%%%%%%%%%%%%%%%%
\subsection{Execution Time Evaluation}
\label{subsec:ExecutionTimeEvaluation}
\begin{figure}[tb]%[htbp]
\begin{center}
\includegraphics[width=0.8\columnwidth]{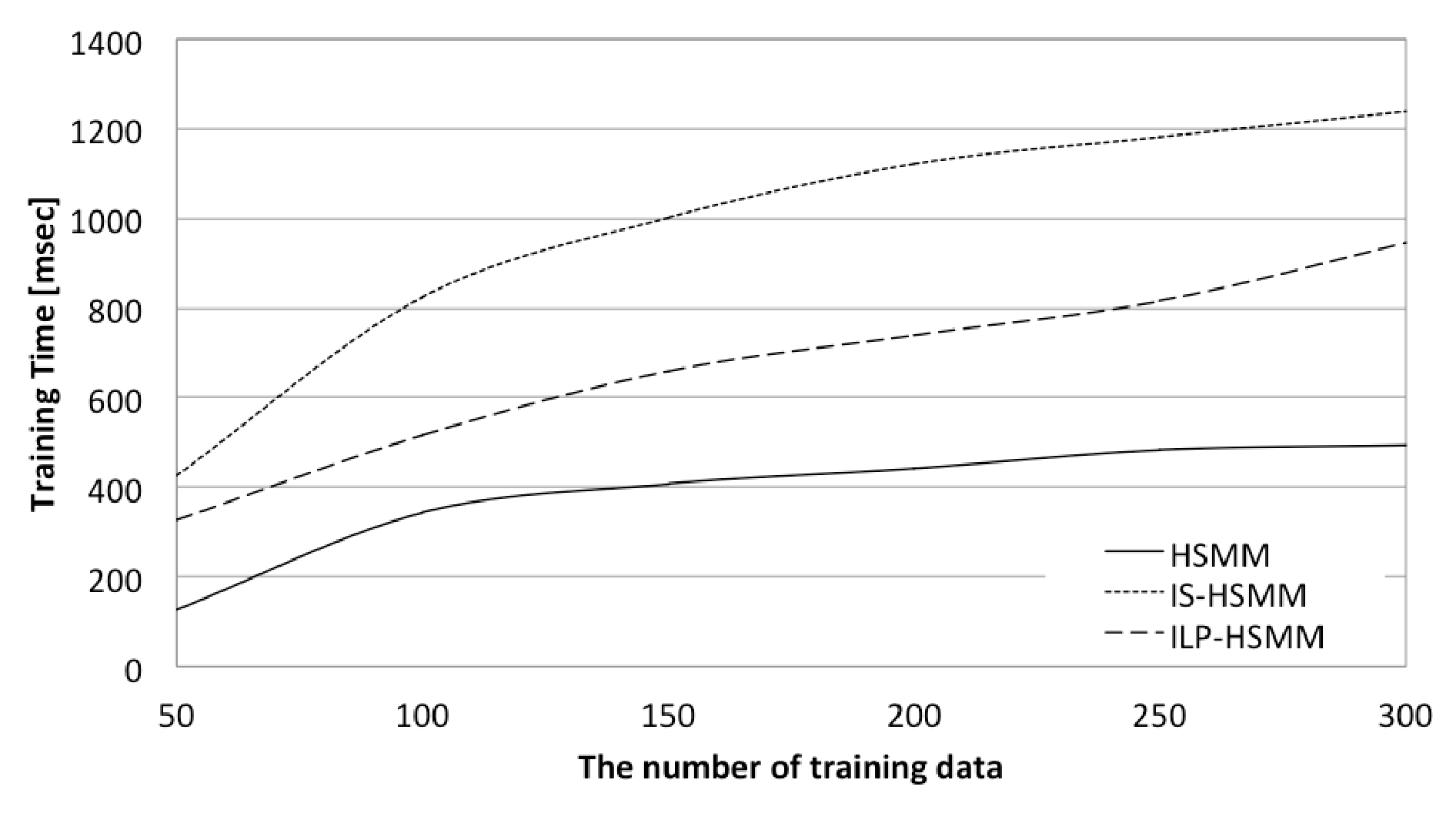}
\caption{Execution time for training.}
\label{fig:TrainingTime}
\end{center}
\end{figure}

This section presents the execution time evaluation for training and recognition. 
For the evaluation, we generate 35 sequences, fixing $d_{min} = d_{max} =2$, $l_{min} =1$, {and} $l_{max} =10$, {where} $T$ is not fixed {\it a priori}. Using the generated data, we compare {the} training time and recognition time while changing the number of training data. The training time results are {presented} in \Figref{fig:TrainingTime}. The $x$-axis shows the number of training data{. T}he $y$-axis shows the execution time for training. The upper, middle, and bottom lines {respectively} present the results of IS-HSMM, ILP-HSMM, and HSMM.
Results show that three graphs are mostly increasing parallel, which shows that the difference between the results of HSMM and IS-HSMM, and the difference between the results of HSMM and ILP-HSMM are both of a certain degree. Therefore, the training time of IS-HSMM and ILP-HSMM requires additional time, but the amount of the additional time does not increase exponentially. 

\begin{figure}[tb]%[htbp]
\begin{center}
\includegraphics[width=0.8\columnwidth]{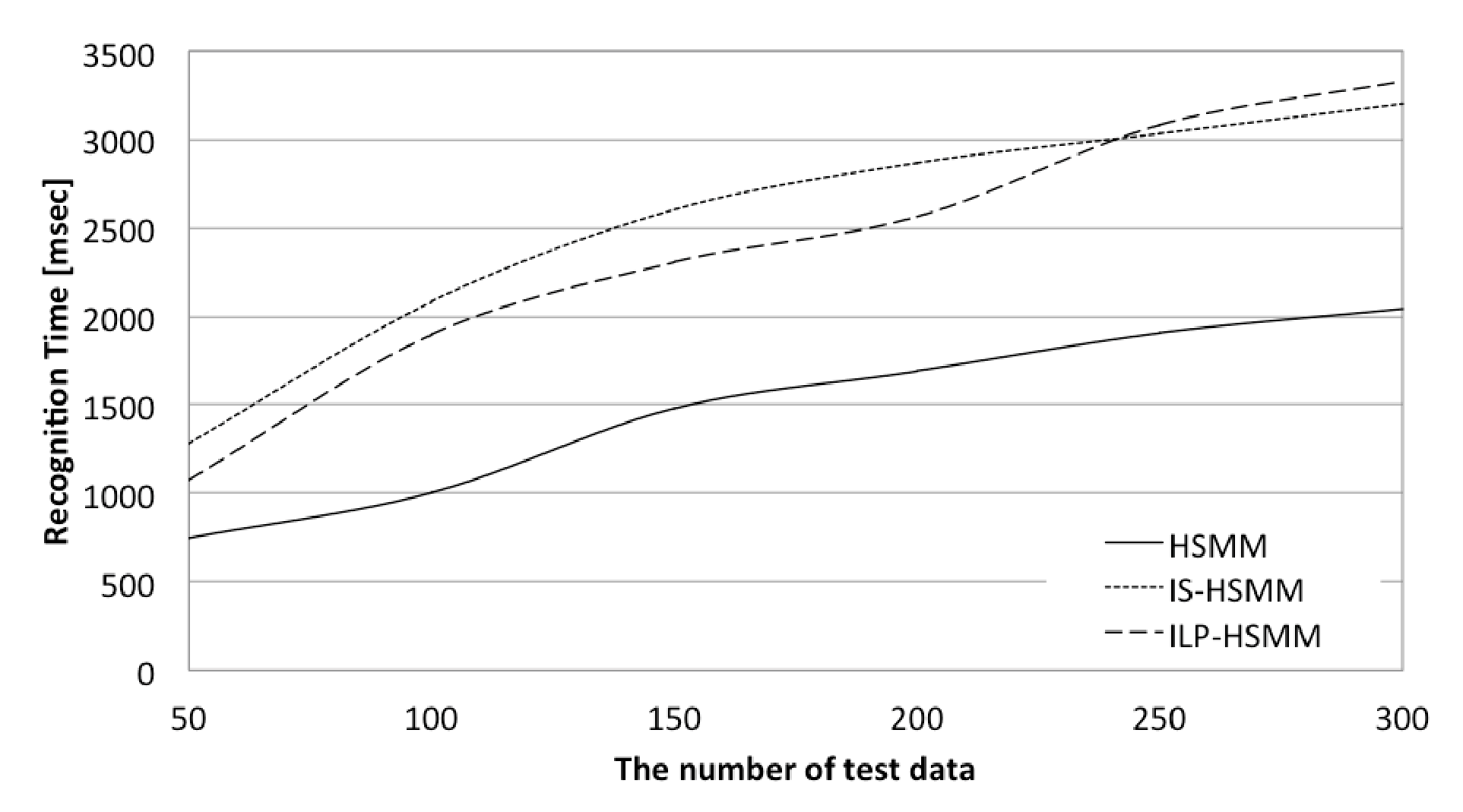}
\caption{Execution time for recognition.}
\label{fig:RecognitionTime}
\end{center}
\end{figure}

Similarly, the execution time for recognition is shown in \Figref{fig:RecognitionTime}. The $x$-axis shows the number of test data{. T}he $y$-axis shows the execution time for recognition. The upper, middle, and bottom lines {respectively} present the results of IS-HSMM, ILP-HSMM, and HSMM. Results {show} that the amount of the additional time for recognition does not increase exponentially to the same degree as training. 
Stated differently, both {the} evaluation results of training time and recognition time reveal that it {causes no severe difficulty} for the execution times. 
%
%
%%%%%%%%%%%%%%%%%%%%%%%
% Figure 12
%%%%%%%%%%%%%%%%%%%%%%%
%
%%%%%%%%%%%%%%%%%%%%%%%
%
%
%%%%%%%%%%%%%%%%%%%%%%%
% Figure 13
%%%%%%%%%%%%%%%%%%%%%%%
%
%%%%%%%%%%%%%%%%%%%%%%%
%
%
%
%%%%%%%%%%%%%%%%%%%%%%%%%%%%%%%%%%
% 
% VIII-C.	Recognition Performance Evaluation    %
%%%%%%%%%%%%%%%%%%%%%%%%%%%%%%%%%%
\subsection{Recognition Performance Evaluation}
\label{subsec:RecognitionPerformanceEvaluation}
This section presents the evaluation results of recognition performance comparing IS-HSMM and ILP-HSMM with HSMM. The evaluation metric is the recognition accuracy based on {the} {\it f}-measure calculated using \\
${{\it f}-measure~=~(2 \cdot {\rm recall} \cdot {\rm precision}) / ({\rm recall} + {\rm precision})}$, where ${\rm precision}=TP/PP$, and ${\rm recall}=TP/AP$. 
Here, the Predicted Positive ($PP$) is the number of models with likelihood calculated using \Eqref{eq:eq8} is maximum in all models{.} True Positive ($TP$) is the number of collected models in $PP$. Actually Positive ($AP$) is the number of labeled models.

\begin{figure}[htb]%[htbp]
\begin{center}
\includegraphics[width=0.8\columnwidth]{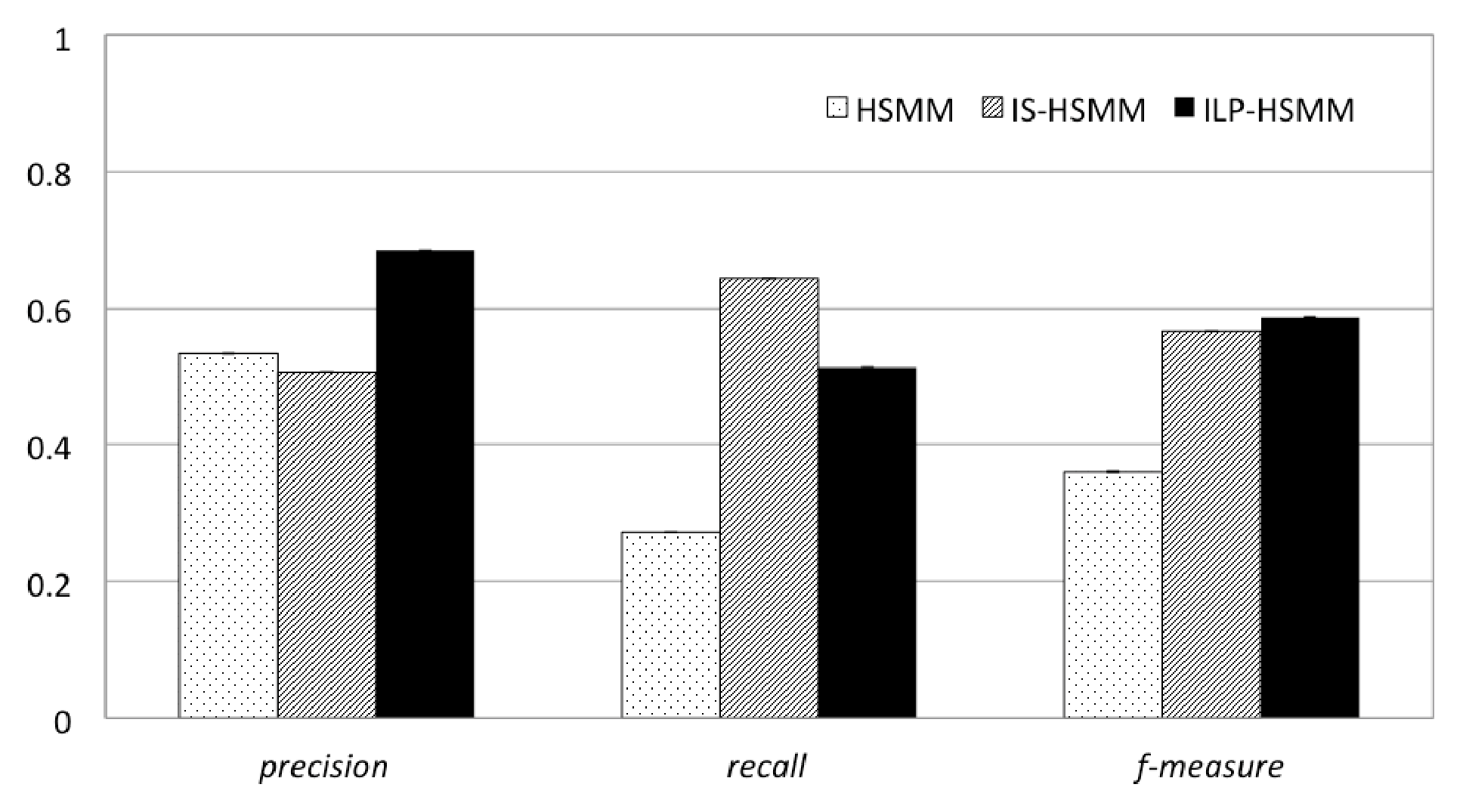}
\caption{Recognition performance: the number of states is 5.}
\label{fig:RecognitionResult}
\end{center}
\end{figure}

\begin{figure}[htb]%[htbp]
\begin{center}
\includegraphics[width=0.8\columnwidth]{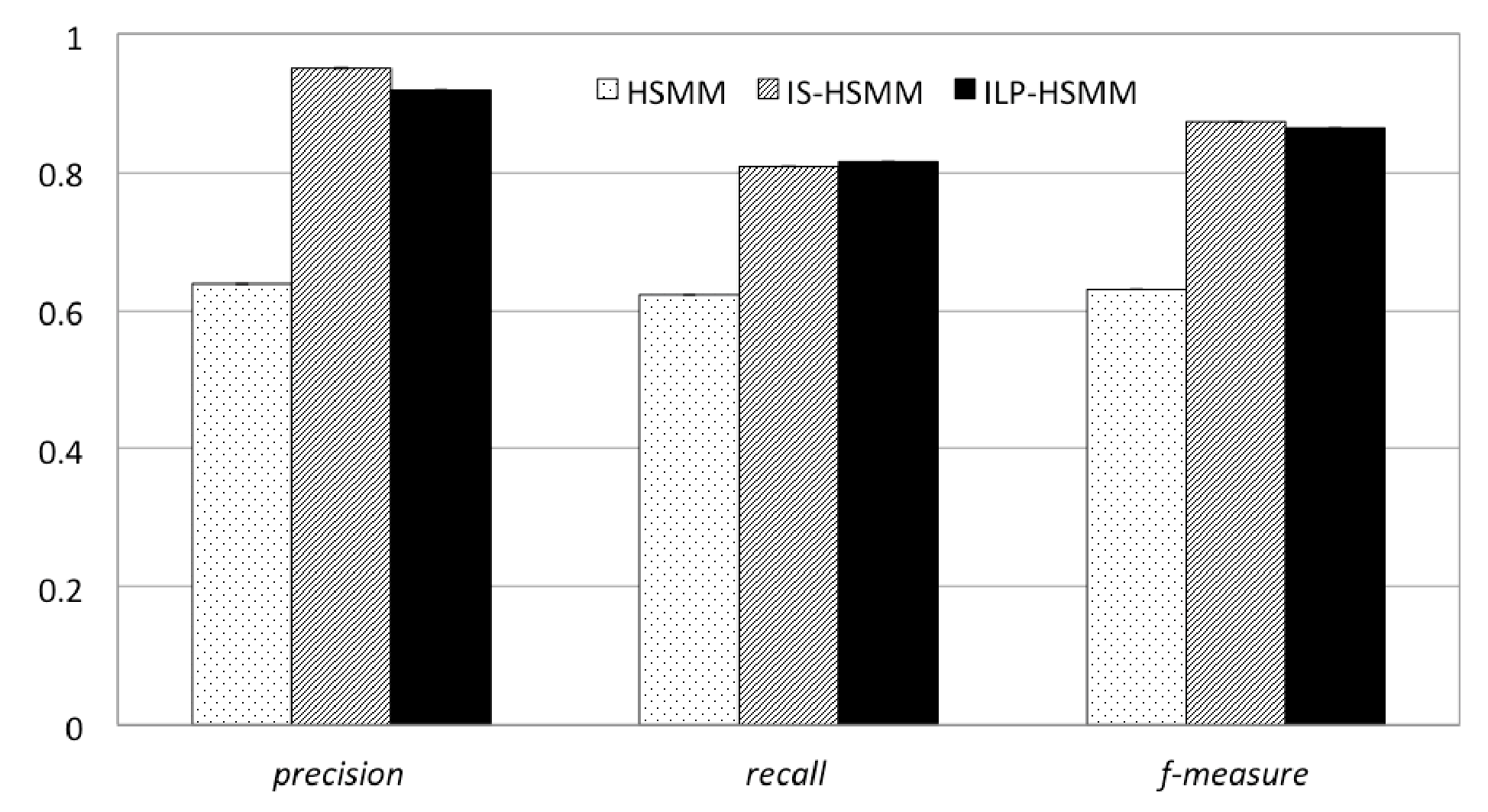}
\caption{Recognition performance: the number of states is 10.}
\label{fig:RecognitionResultState10}
\end{center}
\end{figure}

Results are presented in \Figref{fig:RecognitionResult} and \Figref{fig:RecognitionResultState10}. The $x$-axis {shows} Precision, Recall, and {\it f}-measure{. The} $y$-axis {shows} the score. The left, middle{,} and right bars {respectively} present the results of HSMM, IS-HSMM, and ILP-HSMM. \Figref{fig:RecognitionResult} shows the results obtained when the number of states is {five}, and \Figref{fig:RecognitionResultState10} presents the results obtained when the number of states is {ten}. Both results are the average scores of five repetitions. The results show that both the proposed models IS-HSMM and ILP-HSMM have higher recognition performance than HSMM. By comparing the results of IS-HSMM and ILP-HSMM, the scores of {\it f}-measure are similar, but the scores of recall and precision differ. IS-HSMM has a higher score for recall, but it has lower score for precision than ILP-HSMM. {T}he next section {presents detailed analysis of} the performance{s} of IS-HSMM and ILP-HSMM. 
Finally, comparison of the two results obtained when the numbers of states are {five} and {ten} shows that the recognition performance can be higher depending on the number of states increasing.
%
%%%%%%%%%%%%%%%%%%%%%%%
%Figure 14 (Recognition Performance State = 5)
%%%%%%%%%%%%%%%%%%%%%%%
%
%%%%%%%%%%%%%%%%%%%%%%%
%
%
%
%%%%%%%%%%%%%%%%%%%%%%%
%Figure 15 (Recognition Performance State = 10)
%%%%%%%%%%%%%%%%%%%%%%%
%
%%%%%%%%%%%%%%%%%%%%%%%
%
%
%

\begin{figure}[tb]%[htbp]
\begin{center}
\includegraphics[width=\columnwidth]{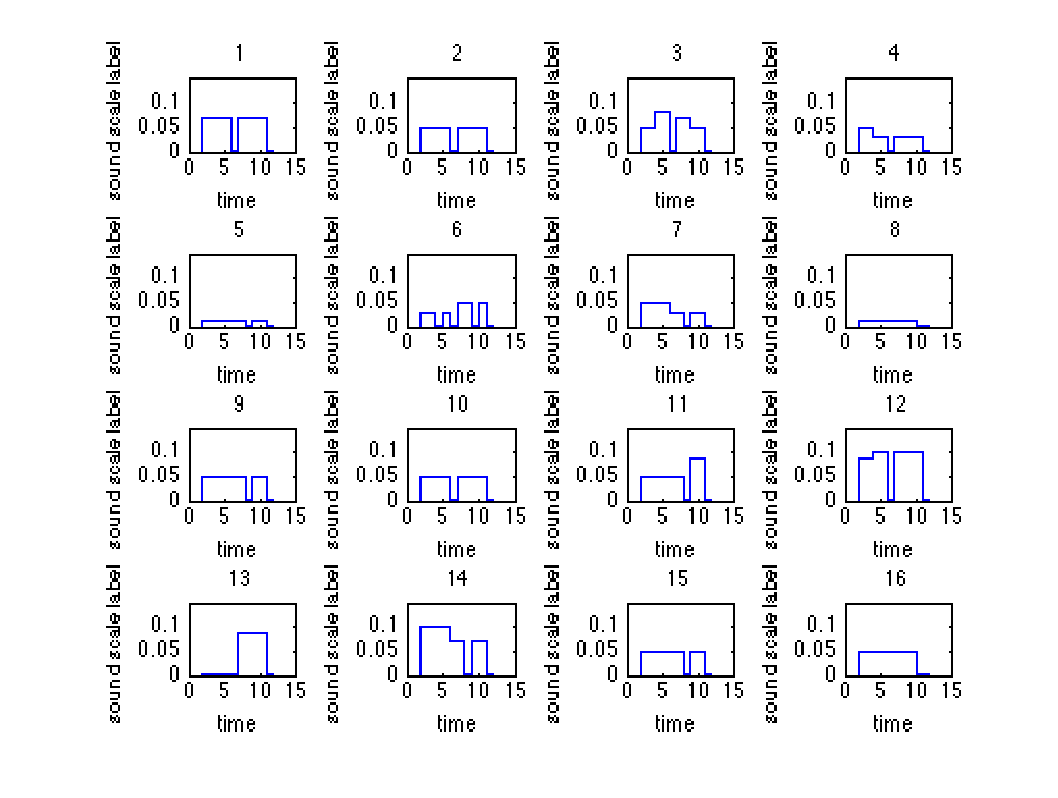} %png
\caption{Musical scale of example input sequences and {their} label{s}.}
\label{fig:dataMusicalScaleStairs}
\end{center}
\end{figure}

\begin{figure}[htb]%[htbp]
\begin{center}
\includegraphics[width=0.8\columnwidth]{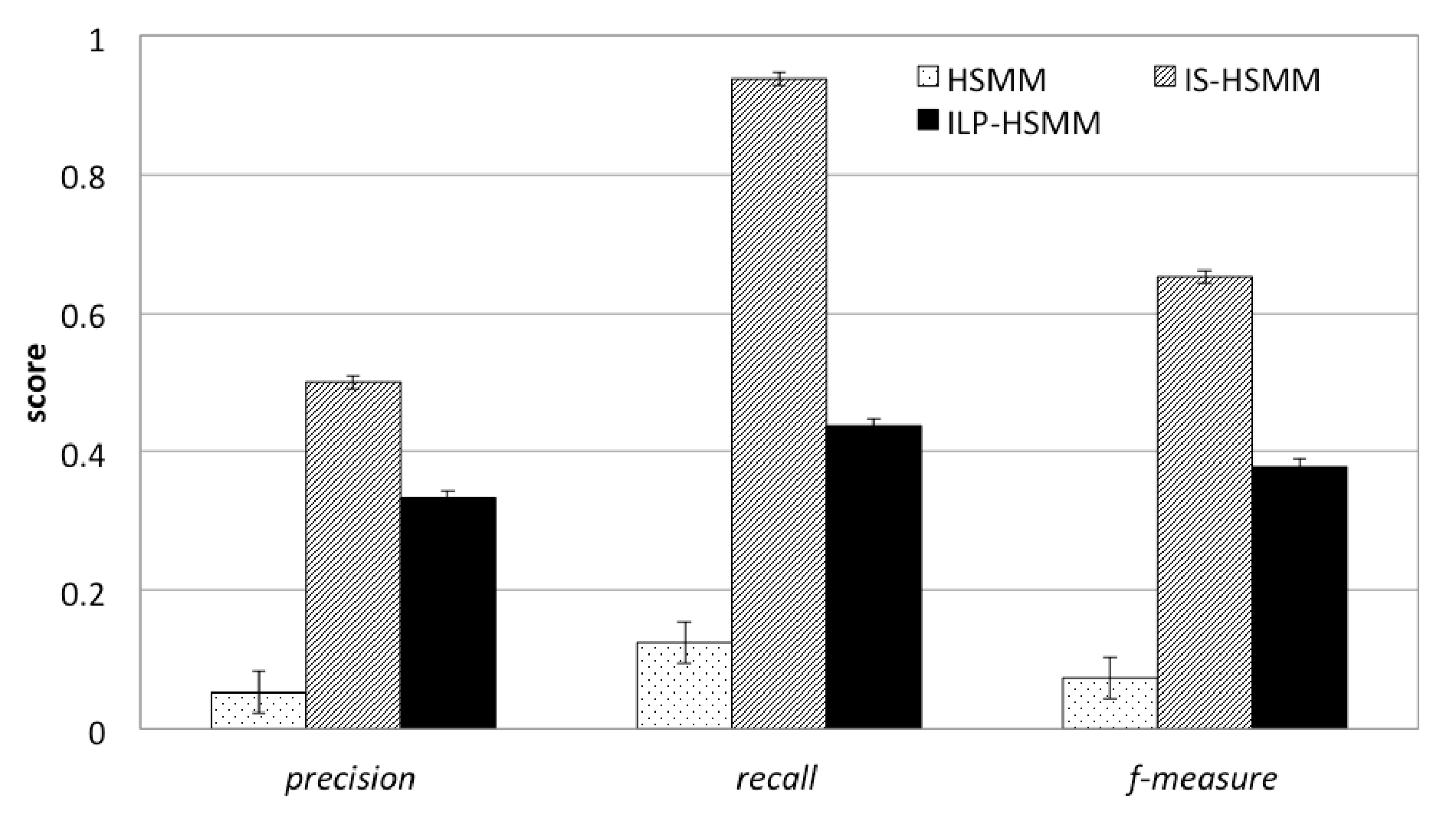}
\caption{Recognition performance with music scale label: the number of states is 2.}
\label{fig:RecognitionResultWithMusicScaleState2}
\end{center}
\end{figure}

\begin{figure}[htb]%[htbp]
\begin{center}
\includegraphics[width=0.8\columnwidth]{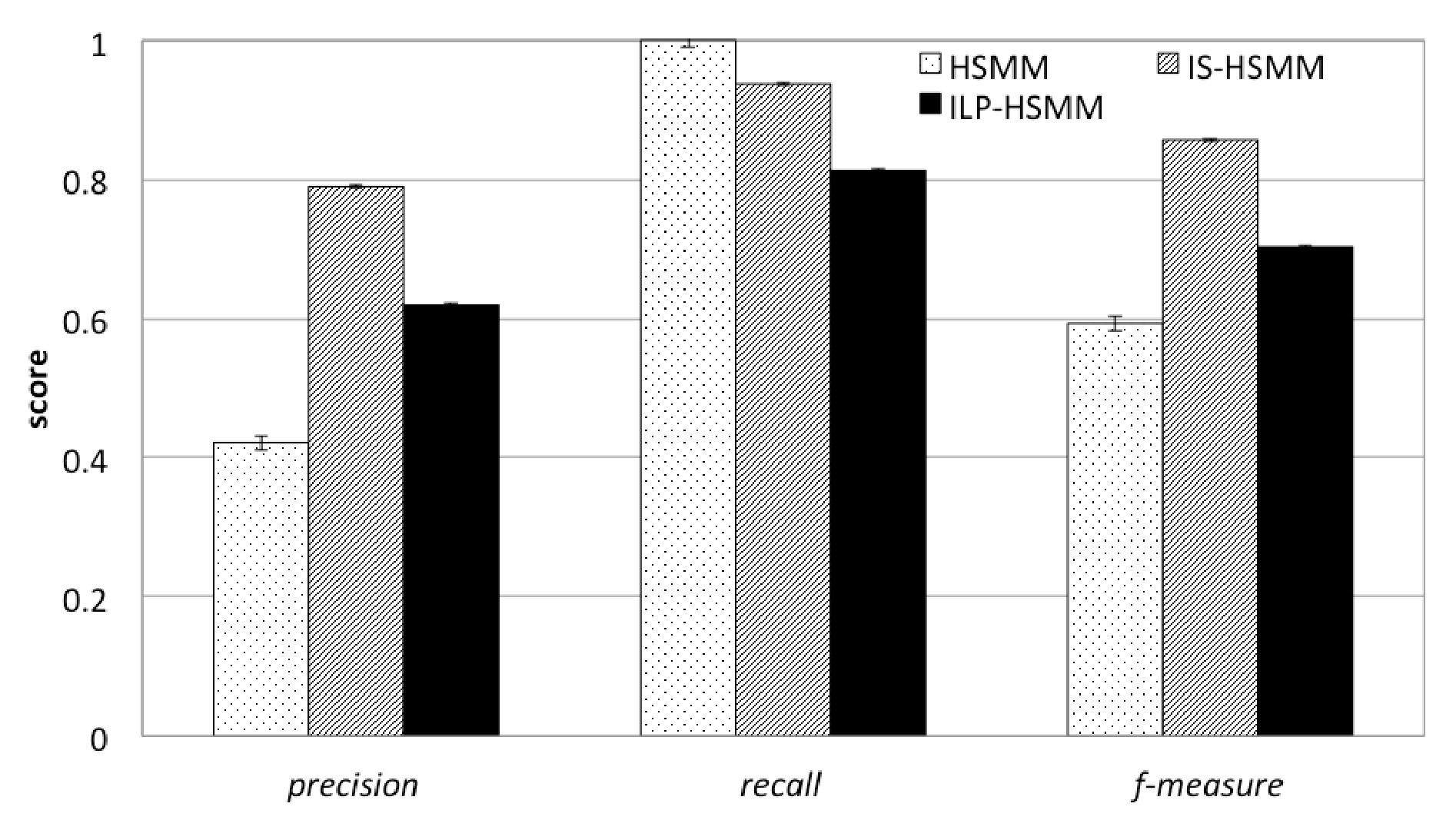}
\caption{Recognition performance with music scale label: the number of states is 10.}
\label{fig:RecognitionResultWithMusicScaleState10}
\end{center}
\end{figure}

The earlier experiment includes observation symbols of only three kinds. To evaluate the performance of treating various durations and intervals with observation symbols of many kinds, we use the musical scale {instead} of the volume of the sound as observation symbols. 
\Figref{fig:dataMusicalScaleStairs} shows the musical scale with stairs of example data. These are the some input data extracted from the evaluation data. The {figure on the top of} each graph signifies the label. Each value from 0.01 to 0.12 in 0.01 intervals is assigned to C, C\#, D, D\#, E, F, F\#, G, G\#, A, A\#, B of the musical scale. If the volume is lower than a threshold, then the value of {the} sound scale label is zero. {This is} the {\it interval observation} in a sequence. 
The results of recognition performance using the data generated as described above are shown in \Figref{fig:RecognitionResultWithMusicScaleState2} and \Figref{fig:RecognitionResultWithMusicScaleState10}. They {present} results of recognition performance evaluation when the numbers of states are 2 and 10. The scores are the average scores of five repetitions. Considering that it would be high performance when the number of states is greater than the number of observation symbols in HSMM, we assign 2 and 10 as the number{s} of states in the experience to compare {their performance}. 

%
%
%%%%%%%%%%%%%%%%%%%%%%%
%Figure (16)
%%%%%%%%%%%%%%%%%%%%%%%
%
%%%%%%%%%%%%%%%%%%%%%%%
%
%

When the number of states is 2, the recognition performance of HSMM is extremely low, but those of IS-HSMM and ILP-HSMM are much higher than HSMM. In addition, the results of IS-HSMM are much higher than ILP-HSMM. However, when the number of states is 10, the number of states is greater than the number of the observation symbols. At this time, the entire scores of HSMM, IS-HSMM, and ILP-HSMM are higher than 0.4. For the HSMM, the recall score gives the max score in all models but the precision score {represents} the lowest value. Therefore, the probability for each sequence using HSMM is similar to that of each other sequence. 
Then, {whereas} the average scores of precision, recall, and {\it f}-measure are more than 0.8 in IS-HSMM, the average score is about 0.7 in ILP-HSMM. As a result, when the number of states increases, the scores of IS-HSMM are higher than those of ILP-HSMM because increasing the states contributes to treatment of the transition probability from a state to another state. 
Therefore, IS-HSMM is effective for treating the order of the sequence precisely 
% because the ``interval" is represented with one of states and HMM can model the transition probability between two states. 
{because it can model the transition probability between two states as the original HMM and it can represent ``interval" as one of the states.}
%by introducing ``interval state" special for representing ``interval"}.
 
However, regarding the input data shown in \Figref{fig:dataMusicalScaleStairs} in detail, No. 4 input data are similar to No. 7; the No. 2 input data are similar to No. 10. It is difficult to distinguish the small time difference between two sequences with both IS-HSMM and ILP-HSMM even if the number of states increases. 
{This difficulty} might cause {a} decline of recognition performance.

Moreover, ILP-HSMM treats the state interval using the new additional parameter between two stationary states. If {the} state interval is mostly similar between static two states, {then} ILP-HSMM can model the length of {the} interval precisely, but it is difficult to model a sequence including various length{s} of duration{s} and interval{s}. 
Therefore, to treat sequential data of various kinds with duration{s} and interval{s}, IS-HSMM would {engender} higher performance than ILP-HSMM. The following section {presents} evaluation results of modeling performance and analysis between ILP-HSMM and IS-HSMM.
% 
% 
% 
% 
% 
%%%%%%%%%%%%%%%%%%%%%%%
%Figure (17) (Recognition Performance with Music Scale State = 2)
%%%%%%%%%%%%%%%%%%%%%%%
%
%%%%%%%%%%%%%%%%%%%%%%%
%
%
%%%%%%%%%%%%%%%%%%%%%%%
%Figure (18) (Recognition Performance with Music Scale State = 10)
%%%%%%%%%%%%%%%%%%%%%%%
%
%%%%%%%%%%%%%%%%%%%%%%%
%
% 
% \clearpage
%
%
%
%
%%%%%%%%%%%%%%%%%%%%%%%%%%%%%%%%%%
% 
%VIII-D.	Performance of Reproducibility
%%%%%%%%%%%%%%%%%%%%%%%%%%%%%%%%%%
\subsection{Reproducibility Performance Evaluations between IS-HSMM and ILP-HSMM} 
\label{subsec:EffectivenessOfProposals}
This section presents the evaluation results of modeling performance, particularly addressing the performance of reproducibility. 
We calculate the performance of reproducibility and compare both IS-HSMM and ILP-HSMM. The {performance of reproducibility} {signifies} how precisely the model generates the original sequence, which is represented as $r$. The $r$ is calculated as
%
%%%%%%%%%%%%%%%%%%%%%%%
%Equation18
%%%%%%%%%%%%%%%%%%%%%%%
%
\begin{eqnarray}\label{eq:eq17}
\hspace{5mm}
r & = &\frac {\sum_{t=1}^{T}(w_t=o_t)} {T},
\nonumber
\end{eqnarray}
%
%%%%%%%%%%%%%%%%%%%%%%%
% 
where $\vec{o}_{1:T}$ {stands for} the original sequence, $T$ {represents} the time length of the original sequence, and $w_{1:T}$ {denotes} the generated sequence using the model parameter $\theta$ which is calculated using the original sequence.
To calculate the equation presented above, we give the sequence length $T$ and generate a sequence which has high likelihood using the forward algorithm with the set of parameters $\mathrm{\Lambda}$. The generated sequence is the estimated sequence. Therefore, the performance of reproducibility
indicates how precisely the model, i.e., the set of parameters $\mathrm{\Lambda}$ 
decided by the training phase, generates the original sequence.

%%%%%%%%%%%%%%%%%%%%%%%
%Figure 19   %% 
%%%%%%%%%%%%%%%%%%%%%%%
%
\begin{figure}[b]%[htbp]
\begin{center}
\includegraphics[width=0.8\columnwidth]{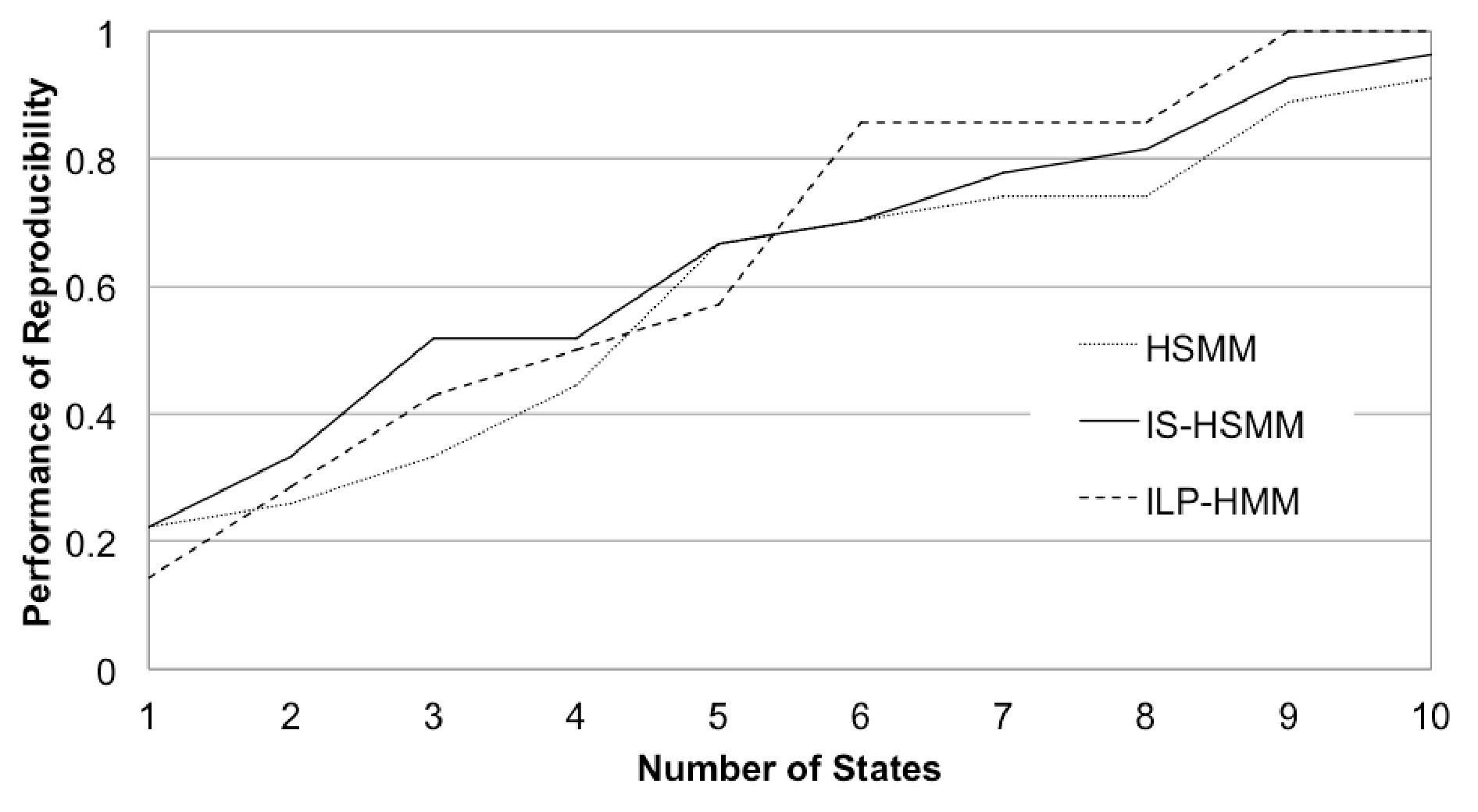}
\caption{{Performance of reproducibility when the number of states increases.}}
%\caption{Modeling performance when the number of intervals increases.}
\label{fig:ModelingPerformanceWithNumOfStates03}
\end{center}
\end{figure}
%%%%%%%%%%%%%%%%%%%%%%%

First, we evaluate the performance of reproducibility when the number of states changes. 
\Figref{fig:ModelingPerformanceWithNumOfStates03} 
presents the results of evaluating reproducibility using HSMM, IS-HSMM{,} and ILP-HSMM. The $x$-axis {shows} the number of states{. T}he $y$-axis {shows} the performance of reproducibility. The number of observed symbols in sequence $N$ is $N=7$. 

Results show that %{all} models obtain higher performance of reproducibility 
{the performance of reproducibility of all models rises} 
{as} the number of states increases. 
The performance results of IS-HSMM and HSMM is mostly the same and IS-HSMM has a bit higher performance than that of HSMM. The results of ILP-HSMM show less performance when {the states are fewer than six}. {They} show higher performance when the number of states is {greater than} six {i.e., the number of observed symbols}. It represents that the number of states is more than the number of observed symbols{;} ILP-HSMM has higher performance of reproducibility than other models.

\begin{figure}[tb]%[htbp]
\begin{center}
\includegraphics[width=0.8\columnwidth]{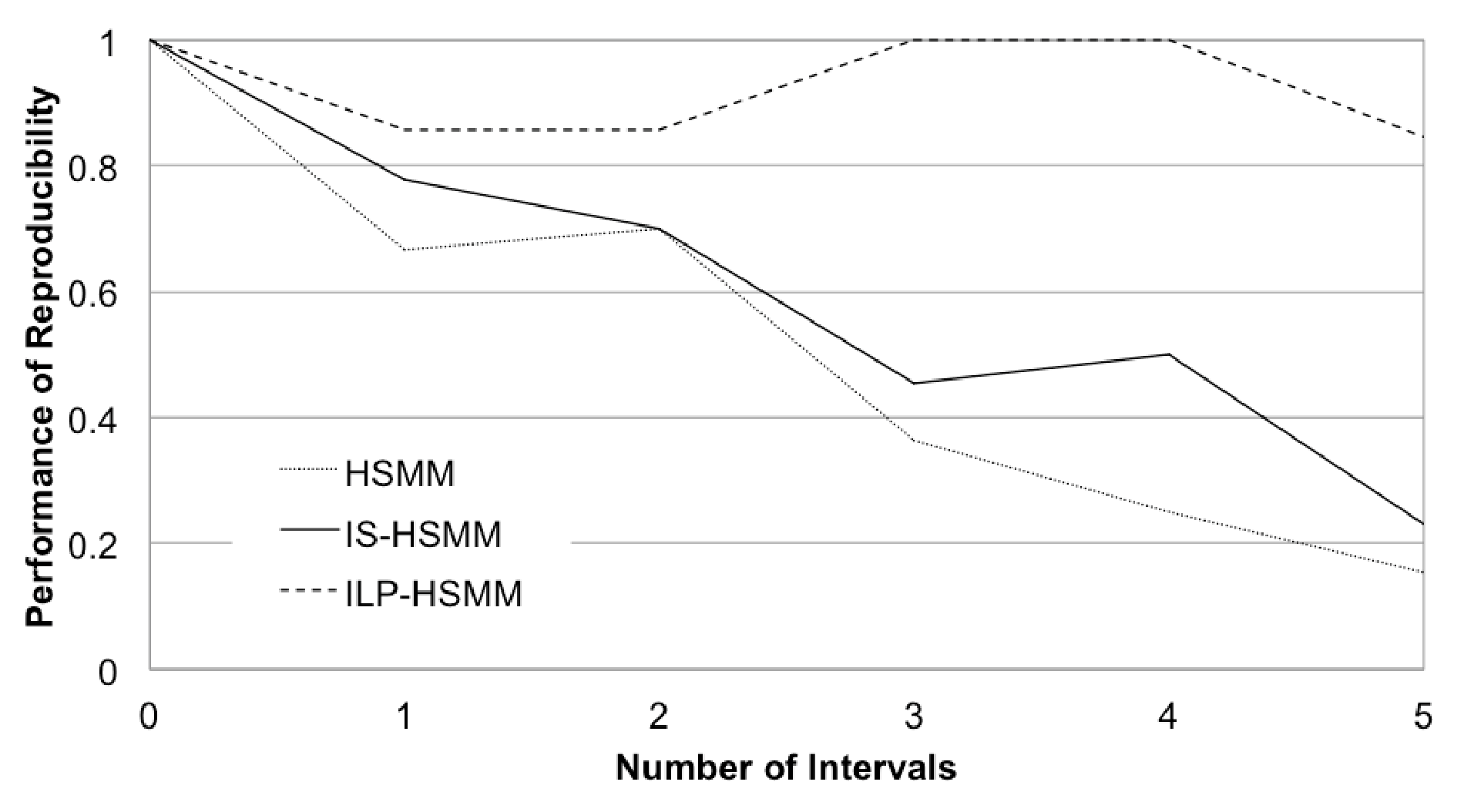}
\caption{{Performance of reproducibility when the number of intervals increases.}}
%\caption{Modeling performance when the intervals become more numerous.}
\label{fig:ModelingPerformanceWithNumOfIntervals03}
\end{center}
\end{figure}

Then, we evaluate the performance of reproducibility when the number of intervals in a sequence changes. 
\Figref{fig:ModelingPerformanceWithNumOfIntervals03} also shows the scores of performance of reproducibility of HSMM, IS-HSMM and ILP-HSMM. 
The $x$-axis {shows} the number of intervals in a sequence{. T}he $y$-axis {shows} the score of {performance of reproducibility}. The number of sorts observed in a sequence is $N=6$. One of the sorts is an interval. 
Results show that the performance of reproducibility of both models; HSMM and IS-HSMM decrease as the number of intervals increases, but 
that of IS-HSMM is higher than that of HSMM. Then, the results of ILP-HSMM is the highest performance in all models. {It} can obtain the highest performance {irrespective} of the number of intervals. 
Therefore, IS-HSMM can model the sequence with intervals more precisely than HSMM{. The} ILP-HSMM can model it most precisely {of} all models.
Comparing two results {of} HSMM and IS-HSMM ensures that the proposed IS-HSMM can model the sequential data more precisely than HSMM by introducing the special state, i.e., the interval state and calculating the transition probability from the state {before} the interval state. In addition, the performance of IS-HSMM is much higher especially when the states are few and even if many intervals exist in a sequence. 
Comparing the other results {for} IS-HSMM and ILP-HSMM ensures that the proposed ILP-HSMM can model the sequential data more precisely than other models because it represents the length of intervals directly in the model. 
%
% 
%
%%%%%%%%%%%%%%%%%%%%%%%
%Figure 20
%%%%%%%%%%%%%%%%%%%%%%%
%
%%%%%%%%%%%%%%%%%%%%%%%

As a result of {the} evaluation {presented} above, both the proposed extension models for HSMM have higher performance than HSMM, but ILP-HSMM can model the static interval between two states. However, it is more important for modeling the general duration and interval using a model {with} trained multiple data which have the same label. From the {perspective} of modeling generalization, the recognition performance of IS-HSMM has a higher score than other models, especially where the number of sorts of the observation symbols is larger. Therefore, we conclude that IS-HSMM has higher performance for modeling the general sequential data, not only for the data which have a static length of interval, but also for data which have various interval lengths.
%
%
%
%
%
%
%%%%%%%%%%%%%%%%%%%%%%%%%%%%%%%%%%
% 
%IX.	Summary and Future Work
%%%%%%%%%%%%%%%%%%%%%%%%%%%%%%%%%%
\section{Summary and Future Work}
\label{sec:Summary}
The goal of this research was to model sequential data, including state duration and {the} state interval, simultaneously. We specifically examined a hidden semi-Markov model (HSMM) to treat such sequential data, and propose{d} two extended models to treat {a} state interval in a sequence: IS-HSMM and ILP-HSMM. IS-HSMM introduces a special calculation technique to treat {an} interval state, where if the preceding state is an interval state, it models the transition from the second preceding state to the current state simultaneously. {However}, ILP-HSMM uses the Gaussian distribution as a length parameter, and trains with both preceding and subsequent states. Comparisons of recognition performance and elapsed time among IS-HSMM, ILP-HSMM, and HSMM {show that} both of the proposed models give higher performance than HSMM although they need additional calculation costs. 
Comparison results between IS-HSMM and ILP-HSMM in terms of the modeling performance reveal that ILP-HSMM has higher performance than that of IS-HSMM.

As {direction of future research}, we intend to {use} our model to treat such actual sensing data which have a feature of rhythm or timing patterns. {Although} ILP-HSMM has higher performance in the evaluation, the concept of IS-HSMM is simpler than that of ILP-HSMM. Additionally, IS-HSMM can adopt another {difficulty} of analyzing sequential data, except for only treating intervals between states. In case the same state occurs frequently in a sequence, it is difficult to model the original sequence precisely {without an interval}. Therefore, we {must} evaluate the effectiveness of treating the original sequence using other application data, and finally extend the model further.

\clearpage	
\bibliographystyle{IEEEtran}
\bibliography{bmc_article}
      	
\end{document}